\documentclass[10pt, a4paper]{article}

\usepackage[final]{lrec-coling2024} 
\usepackage{times}
\usepackage{latexsym}

\usepackage[T1]{fontenc}

\usepackage[utf8]{inputenc}

\usepackage{microtype}

\usepackage{arydshln} 

\usepackage{booktabs}

\usepackage{multirow}

\usepackage{inconsolata}
\usepackage{algpseudocode}
\usepackage[ruled, vlined, linesnumbered]{algorithm2e}
\usepackage{multicol}  
\usepackage{booktabs}
\usepackage{multirow}
\usepackage{graphicx}
\usepackage{rotating}
\usepackage{booktabs}
\usepackage{tabularx}
\usepackage{arydshln}
\usepackage{amsmath}
\usepackage{enumitem}
\usepackage{caption}
\usepackage{soul}
\usepackage{xcolor}
\usepackage[normalem]{ulem}
\usepackage{subcaption}
\DeclareMathOperator*{\argmax}{arg\,max}

\usepackage{cleveref}
\crefformat{section}{\S#2#1#3} 
\crefformat{subsection}{\S#2#1#3}
\crefformat{subsubsection}{\S#2#1#3}
\usepackage{inconsolata}
\newcommand{\pluseq}{\mathrel{+}=}
\title{A Frustratingly Simple Decoding Method for \\ Neural Text Generation}

\name{Haoran Yang$^{{\spadesuit},*}$\thanks{$^{*}$Most work was done during an internship at Tencent AI Lab.}, Deng Cai$^{\heartsuit,\dag}$\thanks{$^{\dag}$Corresponding author.}, Huayang Li$^{\clubsuit}$, Wei Bi$^{\heartsuit}$, Wai Lam$^{\spadesuit}$, Shuming Shi$^{\heartsuit}$} 

\address{$^{\spadesuit}$The Chinese University of Hong Kong~$^{\heartsuit}$Tencent AI Lab~$^{\clubsuit}$Nara Institute of Science and Technology \\
         \{hryang, wlam\}@se.cuhk.edu.hk, li.huayang.lh6@is.naist.jp\\ 
         \{jcykcai, victoriabi, shumingshi\}@tencent.com
         }

\abstract{
We introduce a frustratingly simple, highly efficient, and surprisingly effective decoding method, termed \textbf{F}rustratingly \textbf{S}imple \textbf{D}ecoding (\textbf{FSD}), for neural text generation. The idea behind FSD is straightforward: We construct an anti-language model (anti-LM) based on previously generated text, which is employed to penalize the future generation of repetitive content. The anti-LM can be implemented as simple as an $n$-gram language model or a vectorized variant. In this way, FSD incurs no additional model parameters and negligible computational overhead (FSD can be as fast as greedy search). Despite its simplicity, FSD is surprisingly effective and generalizes across different datasets, models, and languages. Extensive experiments show that FSD outperforms established strong baselines in terms of generation quality, decoding speed, and universality. The code is available at \url{https://github.com/LHRYANG/FSD}
 \\ \newline  }

\begin{document}

\maketitleabstract

\section{Introduction}
Neural text generation has attracted increasing attention from both academia and industry. The canonical approach factors the generation process in an autoregressive fashion, reducing the generation into a series of next-token predictions conditioned on their preceding sequences. With the development of large language models (LMs) \cite{brown2020language,touvron2023llama,touvron2023llama2}, the estimation of the probability distribution for next-token predictions has become remarkably accurate. However, when it comes to open-ended text generation, such as story generation~\cite{fan-etal-2018-hierarchical} and writing assistance~\cite{shi2022effidit}, perhaps counter-intuitively, searching for the most likely sequences (e.g., greedy search and beam search) often results in low-quality outputs. Concretely, the generations are prone to falling into tedious and repetitive loops, a notorious issue referred to as \textit{neural text degeneration} \cite{Holtzman2020The,xu2022learning,shi2024thorough}.
\begin{figure}[t]
 \centering
  \includegraphics[width=1\linewidth]{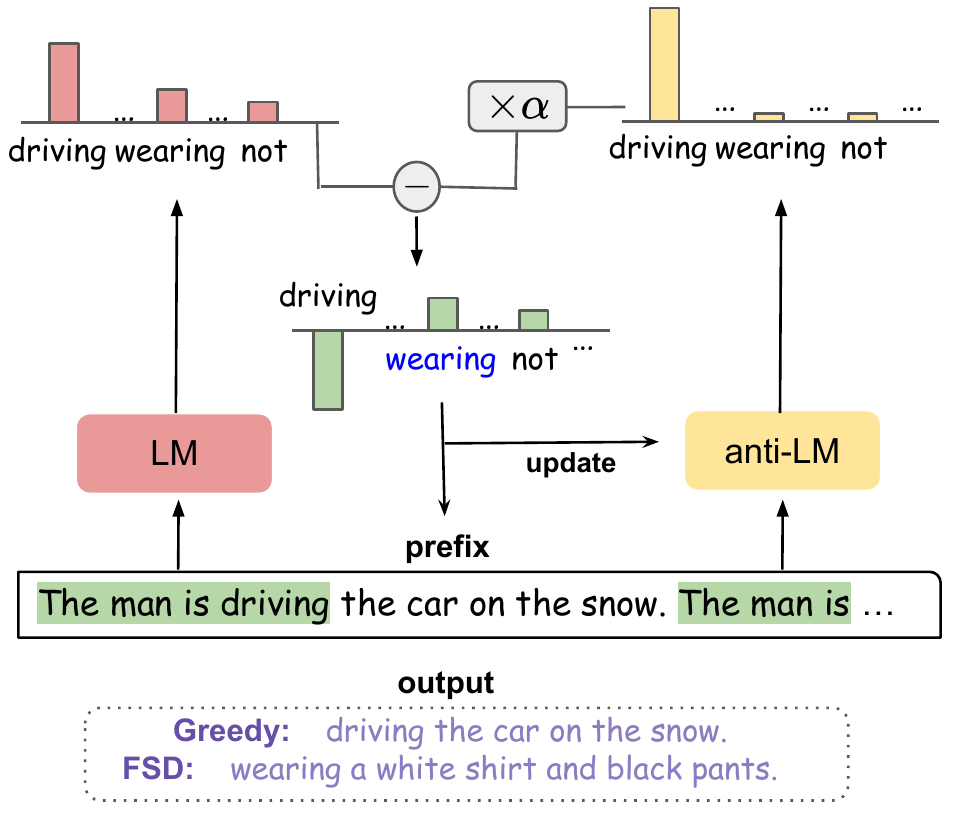}
  \caption{FSD exploits the contrasts between the LM and the anti-LM, where the probabilities from the LM and the anti-LM are used as rewards and penalties respectively. In the above example, the top prediction of the LM is ``driving''. However, the anti-LM also gives a large penalty to ``driving" because it will result in repetition. Consequently, ``wearing'' is instead selected and the anti-LM is updated accordingly.} 
  \label{fig:overview_of_our_model}
\end{figure} 

To address the above problem, two lines of research efforts have been devoted to devising better decoding strategies. The canonical approaches take random samples from the LM's output distribution \cite{fan-etal-2018-hierarchical,Holtzman2020The,meister2022typical,hewitt2022truncation}. The introduced stochasticity can alleviate repetitive generation, however, it also increases the chance of unnatural topic drift and semantic incoherence. More recently, another class of approaches proposes to re-rank top candidate tokens using extra objectives. Concretely, contrastive search (CS)~\cite{su2022a} uses a look-ahead mechanism and penalizes tokens compromising the isotropy of the LM's latent space \cite{ethayarajh2019contextual}. Contrastive decoding (CD)~\cite{li2022contrastive} searches for the token that maximizes the probability difference between the LM and another smaller LM with the same tokenization. Despite better generation quality is achieved, the look-ahead mechanism in CS and the running of an external LM in CD considerably increase computational overhead. Moreover, CS relies on the isotropic property of the LM and CD depends on another LM using the same tokenization, thereby limiting their applicability.

In this paper, we propose \textbf{F}rustratingly \textbf{S}imple \textbf{D}ecoding (FSD) for addressing the degeneration issue with minimal computational cost and without any assumptions on the underlying LM. As illustrated in Figure~\ref{fig:overview_of_our_model}, FSD works by imposing penalties on repetitive patterns that have appeared in the prefix. This is realized through an anti-LM that can capture and memorize these patterns. Specifically, at each generation step, both the LM and the anti-LM take the current prefix as input and separately produce two next-token distributions. The generation probabilities from the LM serve as rewards and those from the anti-LM act as penalties. FSD subtracts the penalties from the rewards, selects the token that maximizes the final score, and continuously updates the anti-LM based on the growing prefix. The anti-LM can be implemented as simple as an $n$-gram language model or a vectorized variant, making FSD as fast as greedy search.

We perform extensive experiments to demonstrate the effectiveness, efficiency, and universality of FSD. The key findings can be summarized as follows: (1) On three canonical open-ended text generation benchmarks, the generation quality of FSD not only surpasses the standard top-$p$ sampling but also is comparable to, if not better than, recent state-of-the-art methods, according to both automatic and human evaluations. (2) FSD exhibits robustness in handling varying generation lengths, particularly demonstrating its superiority in generating longer sequences where existing state-of-the-art methods often struggle. (3) The generation speed of FSD is as fast as greedy search (the theoretical upper bound for autoregressive generation). The speed advantage over existing state-of-the-art methods amplifies as the generation length increases. (4) FSD shows versatility across a variety of models, languages, and tasks (e.g., instruction following and summarization).

\section{Related Work}
Recent years have witnessed enormous progress in neural text generation, particularly with the success of large LMs \cite{radford2019language}. The most straightforward heuristics for generating text from an LM is to find the most likely sequence estimated by the LM. Although maximizing the LM probabilities (e.g., greedy search and beam search) obtains excellent performance in close-ended text generation tasks (e.g., translation~\cite{sutskever2014sequence} and summarization~\cite{see-etal-2017-get}), these search-based methods suffer from generating nonsensical output in open-ended text generation tasks (e.g., story generation \cite{fan-etal-2018-hierarchical}). One prominent issue is that they tend to generate dull and repetitive output \cite{Holtzman2020The,fu2020a,pillutla2021mauve}.

\paragraph{Decoding Methods}
To tackle the above challenge, different decoding methods have been proposed, which can be broadly categorized into two classes. The first class is truncated sampling, where each token is randomly sampled from a truncated next-token distribution. For instance, top-$k$ sampling~\cite{fan-etal-2018-hierarchical} only samples from the $k$ most likely tokens. Top-$p$ sampling~\cite{Holtzman2020The} only considers the minimal set of top tokens that cover a specified percentage $p$ of the distribution. Typical sampling ~\cite{meister2022typical} sorts tokens according to the differences between distribution entropy and probabilities. \citet{hewitt2022truncation} truncate words whose probabilities are below an entropy-dependent threshold. Although sampling-based methods reduce repetitions, the randomness at each sampling step also increases the chance of incoherence and topic drift.

The second class of decoding methods is still search-based but optimizes a different objective. Contrastive Search (CS)~\cite{su2022a} assumes the LM has an isotropic representation space and adds a penalty term that decreases the generation probabilities of tokens producing hidden states that are similar to the previous context. However, the look-ahead operation at each step brings considerable additional cost. Contrastive Decoding (CD)~\cite{li2022contrastive} employs an amateur LM (a smaller pre-trained LM using the same tokenization) and penalizes undesired attributes associated with the amateur model. In contrast, FSD is much more lightweight and efficient; FSD only constructs an $n$-gram model on-the-fly, requiring no external model and introducing negligible computational cost. In addition, FSD holds the potential for broader applicability as it does not assume the existence of an amateur LM or any properties of the LM.

\paragraph{Training Methods}
Another group of methods attempts to improve text generation quality by fine-tuning the LMs with new training objectives. \citet{Welleck2020Neural} propose unlikelihood training, which explicitly minimizes the generation probability of repetitive tokens. \citet{lagutin-etal-2021-implicit} improve the generation using policy gradient with a repetition objective.
\citet{xu2022learning} learn to penalize probabilities of sentence-level repetitions from pseudo-repetitive data. \citet{su2022a} devise a contrastive training objective that encourages discriminative and isotropic token representations. In contrast, FSD simply employs off-the-shelf pre-trained LMs and requires zero training.
\section{Background}

\subsection{Language Models}
An LM is a probability distribution over token sequences. Given a sequence $x_{1:t} = x_1, x_2, \ldots, x_{t}$ of length $t$, LM assigns a probability $p(x_{1:t})$ to the sequence, which is usually decomposed in an autoregressive fashion: $p(x_{1:t})= \prod_{i=1}^{t} p(x_i|x_{<i})$.
\paragraph{$N$-gram Language Model}
The most traditional LM is the $n$-gram model, which relies on the Markov assumption ~\cite{Jurafsky2009}. In an $n$-gram LM, the probability of the $i$-th token only depends on the previous $n-1$ tokens, expressed as $p(x_i|x_{<i})= p_n(x_i|x_{i-n+1:i-1})$. This probability can be computed by evaluating the relative frequency counts within a training corpus:
\begin{equation}
\label{eq:ngram}
    p_n(x_{i}|x_{i-n+1:i-1})= \frac{C(x_{i-n+1: i})}{C(x_{i-n+1:i-1})}
\end{equation}
where $C(\cdot)$ counts the number of occurrences of the input sequence within the training corpus. In practice, the probability distributions are often smoothed to improve the model's generalizability. For example, the interpolation of $n$-gram models of different orders can help prevent the LM from assigning zero probability to unseen sequences \cite{inter_ngram}.

\paragraph{Neural Language Model}
With the rise of deep learning, $n$-gram LMs have been largely superseded by neural networks, for example, the GPT family~\cite{radford2019language,brown2020language} and the LLaMA family \cite{touvron2023llama,touvron2023llama2}. These models are trained to predict the next token by conditioning on the preceding context: $\mathcal{L}_{\theta} = -\sum_{i=1}^{t} \log p_\theta (x_i|x_{<i})$, where $\theta$ denotes the model parameters. With the capabilities acquired by large-scale pre-training, these neural LMs can be readily applied to text generation~\cite{prompt}.

\subsection{Open-Ended Text Generation}
Most of our experiments are conducted on open-ended text generation tasks, where the input is a short prompt and the goal is to generate a fluent and coherent continuation. Formally, given a prompt $x_{1:l} = x_1, x_2, \ldots, x_l$, we aim to generate the next $m$ tokens, denoted by $x_{l+1:l+m}=x_{l+1}, x_{l+2}, \ldots, x_{l+m}$. A pre-trained neural LM can complete this task autoregressively by a series of next-token predictions:
\begin{equation}
    p_\theta(x_{l+1:l+m}|x_{1:l}) = \prod_{i=l+1}^{l+m}p_\theta(x_i|x_{<i})\label{eq:mle}
    \nonumber
\end{equation}
Previous works have revealed that the decoding method that selects the token at each generation step has a significant impact on the generation quality \cite{Holtzman2020The,wiher2022decoding}. For example, greedy and beam search often result in repetitions while sampling-based methods suffer from incoherence \cite{su2022a,li2022contrastive}.

\section{Method}
We present our proposed decoding method, Frustratingly Simple Decoding (FSD), named after its remarkably straightforward nature. We begin by introducing the intuition and the general framework of FSD (\cref{sec:intuition}). We then describe the implementation of FSD in the discrete version (\cref{sec:method}) and further extend it to the vectorized version (\cref{sec:vector}).

\subsection{Intuition \& Framework}
\label{sec:intuition}
To produce coherent and diverse generations, it is crucial not only to select the most probable tokens but also to prevent repetitive content. While the former objective can be achieved using the original LM, the latter requires a mechanism for tracking previously generated content and reducing their likelihood of reoccurrence. To this end, we propose the construction of an anti-LM based on the preceding context. This anti-LM is expected to assign higher scores to tokens that will cause repetitions in the preceding context. Consequently, these scores serve as penalties. By integrating the original LM and the anti-LM, we can discourage repetitive token generation and promote other contextually appropriate choices.

Formally, when decoding the $t$-th token, we calculate an $\mathrm{FSD}$ score for each candidate token $v$: 
 \begin{equation}
 \label{eq:fsd}
     \mathrm{FSD}(v|x_{< t}) = p_{\theta}(v|x_{< t})-\alpha \times p_{\omega}(v|x_{< t})
 \end{equation}
where $p_\theta$ and $p_\omega$ represent the LM and the anti-LM respectively. The hyper-parameter $\alpha \geq 0$ is used to balance the two scores. In practice, we first select the top-$k$ most probable tokens according to $p_\theta (\cdot|x)$, denoted by $V^{(k)}$. The token in $V^{(k)}$ with the largest $\mathrm{FSD}$ score is chosen as the $t$-th token.
\subsection{$N$-gram Model as anti-LM}
\label{sec:method}
Following the intuition described above, we start to devise the anti-LM. In principle, any language model capable of capturing patterns in a token sequence can be harnessed to implement the anti-LM. However, we note several critical design principles. First, the prediction of the anti-LM should be efficient, given that it is invoked at every decoding step. Second, the anti-LM should not assume any particular properties of the LM or the language, thus ensuring our method's universal applicability across diverse settings. Last but not least, the update of the anti-LM should be easy, as it undergoes continuous evolution with the expanding prefix.

One natural (and perhaps the simplest) choice is the $n$-gram LM, which offers two key advantages. First, all the operations (i.e., construction, prediction, and update) associated with an $n$-gram model add little computational overhead. Second, the effectiveness and efficiency of $n$-gram LM is scalable across different prefix lengths.
\paragraph{Construction and Update}
Given an input prompt $x_{1:l}$, the $n$-gram LM is constructed and updated as follows. Initially, the prompt $x_{1:l}$ is split into $n$-grams. These $n$-grams can be stored as a set of key-value pairs $\mathcal{D}_n$. For each $n$-gram $x_{i-n+1:i}$, the key is the first $n-1$ tokens $x_{i-n+1:i-1}$ and the value is the last token $x_i$. After generating each new token, we update $\mathcal{D}_n$ to include the new $n$-gram composed by the last $n$ tokens in the sequence.

To calculate next-token probabilities, we use the last $n-1$ tokens in the sequence as the query. We first identify all key-value pairs in $\mathcal{D}_n$ whose key precisely matches the query and then compute the probabilities according to Eq. \ref{eq:ngram}. All of the above operations introduce little computational overhead compared to the running of the original neural LM. 
\begin{algorithm}[t]
   \small
   \SetAlCapFnt{\small}
   \SetAlCapNameFnt{\small}
    \caption{Calculation of Penalty $p_\omega(v|x_{< t})$}
    \SetKwInOut{Input}{Input}
    \SetKwInOut{Output}{Output}
    \label{alg:lambda}
    \Input{prefix $x_{<t}$; $n$-gram models with different orders from 1 to  $N$ ($p_1, p_2, \cdots p_N$); candidate token $v$; decay factor $\beta=0.9$} 
    Initialize $r=1, c_v = 0$\\
    \For{$n=N,\cdots, 2$}
    {
        \If{$p_n(v|x_{< t}) \neq 0$}{
         $\lambda_n = r*\beta$ \\
         $r = r - \lambda_i$ \\
         $c_v \pluseq \lambda_n p_n(v|x_{t-n+1:t-1})$ \\
        }
    }
    $c_v \pluseq r p_1(v)$ \\
    \Output{$ p_\omega(v|x_{< t}) = c_v$}
\end{algorithm}
\paragraph{Smoothed $N$-gram Model}
An ordinary $n$-gram model cannot penalize the $m(m<n)$-gram repetitions. Inspired by two common smoothing techniques in modern $n$-gram models, back-off and interpolation~\cite{Jurafsky2009}, we combine $n$-gram models with different orders from $n=1$ to $N$ ($N$ being the highest order). The result is a smoothed $n$-gram model $\hat{p}$:
\begin{equation}
    \hat{p} = \lambda_{N} p_N + \lambda_{N-1} p_{N-1} + \cdots + \lambda_1 p_1
    \label{eq:interpolation}
\end{equation}
where $\lambda_n$ is the weight of $p_n$ and $\sum_{n=1}^{N}\lambda_n=1$. The detailed process is elaborated in Alg.~\ref{alg:lambda}. In brief, we enumerate $n$-gram models from $n=N$ to $n=1$, setting $\lambda_n$ to decrease exponentially with a decay factor $\beta=0.9$, thus assigning greater weights to higher-order sub-models. The construction and update of the smoothed $n$-gram LM are straightforward; We only need to maintain $N$ copies of key-value pairs ($\mathcal{D}_1, \mathcal{D}_2, \ldots, \mathcal{D}_N$) separately. 

\subsection{Vectorized $N$-gram Model}
\label{sec:vector}

We further provide a vectorized version where the keys are represented using continuous vectors instead of discrete tokens. It offers two advantages compared with the discrete version. First, it possesses the ability to penalize not only identical but also similar patterns in the preceding context, thus allowing for more generalizable pattern recognition. Second, the computation of the vectorized version can be efficiently conducted on GPU, resulting in faster decoding speed.

Specifically, we use the hidden states from the last layer of the original LM as the keys. Let $\boldsymbol{h}_1, \boldsymbol{h}_2, \ldots, \boldsymbol{h}_{t-1}$ be the hidden states for the current sequence $x_{1:t-1}$ ($\boldsymbol{h}_{t-1}$ is used to predict the $t$-th token in the original LM). Each key-value pair in the discrete version $(x_{i-n+1:i-1}, x_i)$ now turns to be $(\boldsymbol{h}_{i-n+1:i-1}, x_i)$. Accordingly, the exact query-key matching in the discrete version becomes a ``soft'' vector matching. To predict the next token, the query is $\boldsymbol{h}_{t-n+1: t-1}$ and the matching score between the query and a key $\boldsymbol{h}_{i-n+1:i-1}$ is computed as follows:
\begin{equation}
\label{eq:vectorized}
    c_i = \mathrm{cos}(\mathrm{cat}(\boldsymbol{h}_{i-n+1:i-1}), \mathrm{cat}(\boldsymbol{h}_{t-n+1:t-1}))
\end{equation}
where $\mathrm{cos}$ computes cosine similarity and $ \mathrm{cat}$ denotes vector concatenation. For a candidate token $v$ that appears multiple times in the sequence, we take the largest matching score as its penalty score. In addition, we clip the penalty score to ensure it is always greater than or equal to zero.
\section{Experiments}
 \begin{table*}[th]
   \small
    \centering
    
    \begin{tabular}{cc|ccc|ccc|ccc}
    \toprule
    & &\multicolumn{3}{c}{wikinews}&\multicolumn{3}{c}{wikitext}&\multicolumn{3}{c}{book} \\
        & &\textbf{div} &\textbf{mau} &\textbf{coh}  &\textbf{div} &\textbf{mau} &\textbf{coh}  &\textbf{div} &\textbf{mau} &\textbf{coh} \\
       \midrule  
       & Human &0.92 &1 &0.65 &0.93 &1 &0.63  &0.95 &1 &0.51 \\

      \midrule 
    \multirow{7}{*}{\begin{sideways} \scriptsize OPT-6.7b\end{sideways}}

    &p=0.95 &\textbf{0.91} &\textbf{0.95} &0.60 &0.87 &\textbf{0.95} &0.59 & 0.93 &0.92 &0.48\\
    &typical=0.95 &0.94 &0.93 &0.58 &\textbf{0.93} &0.93 &0.56 &\textbf{0.95} &0.89 &0.45\\
    &CS  &\textbf{0.91} &0.93 &0.62 &0.87 &0.91 &0.57 &0.86 &0.88 &0.47\\
    &CD  &\textbf{0.93} &\textbf{0.95} &0.69 &0.89 &\textbf{0.95} &0.69 &0.87 &\textbf{0.95} &0.61\\ 
    &FSD &0.95 &0.93 &\textbf{0.66} &0.94 &0.93 &\textbf{0.61} &\textbf{0.95} &0.85 &\textbf{0.51}\\ 
    &FSD-vec &0.95 &0.93 &\textbf{0.64} &0.92 &0.90 &0.60  &0.96 &0.87 &0.49\\

       \midrule 
    \multirow{7}{*}{\begin{sideways} \scriptsize GPT2-XL\end{sideways}}

    &p=0.95 &0.94 &\textbf{0.96} &0.60 &0.92 &\textbf{0.94} &0.57 & \textbf{0.94} &\textbf{0.95} &0.46\\
    &typical=0.95 &0.95 &0.93 &0.56 &0.95 &0.92 &0.53 &\textbf{0.96} &0.87 &0.43\\
    &CS  &0.93 &0.92 &\textbf{0.64} &0.86 &0.92 &0.60 &0.88 &0.89 &0.48\\
    &CD  &\textbf{0.92} &0.92 &0.69 &0.89 &0.93 &0.69 &0.83 &0.93 &0.64\\ 
    &FSD &0.93 &0.93 &\textbf{0.66} &0.94 &0.88 &\textbf{0.61} &\textbf{0.96} &0.90 &\textbf{0.49}\\ 
    &FSD-vec &0.93 &0.93 &\textbf{0.64} &\textbf{0.93} &0.90 &0.58  &\textbf{0.96} &0.91 &0.47\\
    
    
    \midrule 
    \multirow{7}{*}{\begin{sideways} \scriptsize GPT2-Medium\end{sideways}}
    &p=0.95 &0.96 &0.94 &0.56 &0.96 &0.92 &0.53 &\textbf{0.97} &\textbf{0.92} &0.43\\
    &typical=0.95 &0.96 &0.94 &0.56 &0.96 &\textbf{0.93} &0.53 &\textbf{0.97} &0.91 &0.43\\
    &CS& 0.03 &0.14 &\textbf{0.65} &0.02 &0.07 &\textbf{0.64} &0.01 &0.03 &\textbf{0.50}\\
    &CD &0.88 &\textbf{0.95} &0.71 &0.83 &0.88 &0.71 & 0.68 &\textbf{0.92} &0.67\\ 
    &FSD &\textbf{0.94} &0.93 &\textbf{0.65} &\textbf{0.94} &0.91 &0.60 &\textbf{0.97} &0.87 &0.49\\ 
    &FSD-vec &\textbf{0.94} &0.90 &0.60&\textbf{0.92} &0.86 &0.55 &\textbf{0.93} &\textbf{0.92} &0.44\\
   
    \bottomrule
    \end{tabular}
    
    \caption{Automatic evaluation results. The best results (\textbf{the closer to human the better}) are boldfaced.}
    \label{tab:auto}
\end{table*}
Our main experiments focus on open-ended text generation. This task has been used for evaluating various decoding methods in recent works~\cite{li2022contrastive,su2022a,lan2022momentum} because it is particularly susceptible to the repetition issue. We follow the standard setups (\cref{sec:setup}) and report the results in \cref{sec:results}. In addition, we assess the speed of the decoding methods in \cref{sub_sec:speed}, an essential aspect when considering real-world deployment. Moreover, we explore the universality of our proposed method in \cref{sec:uni} from several perspectives: (1) robustness across various models, languages, and datasets (2) versatility for tackling other tasks such as instruction following (the most popular use of LLMs) and closed-ended generation.

\subsection{Setup for Open-Ended Text Generation}
\label{sec:setup} 

\paragraph{Datasets \& Models} 
Following previous works \cite{su2022a,li2022contrastive,lan2022momentum}, we compare FSD and existing decoding methods on three English benchmarks. That is, wikinews\footnote{\url{http://www.wikinews.org}} in the news domain, wikitext-103~\cite{merity2017pointer} in the Wikipedia domain and bookcorpus~\cite{zhu2015aligning} in the story domain. For each test case, the first 32 tokens are used as the prompt and the task is to generate the following 256 tokens. We test three off-the-shelf LMs of different scales: OPT-6.7b~\cite{zhang2022opt}, GPT2-XL, and GPT2-Medium~\cite{radford2019language}. The amateur LM used in CD is OPT-125m for OPT-6.7b and GPT2 for GPT2-XL and GPT2-Medium. 

\paragraph{Evaluation Metrics}

For automatic evaluation, we report three metrics assessing different aspects of the generations:
\begin{itemize}[wide=0.\parindent,noitemsep,topsep=0.em]
    \item \textbf{Diversity} measures the degree of repetition at different $n$-gram levels. The calculation can be expressed as $\prod \limits_{n=2}^{4}(1-\texttt{REP}_n)$, where $\texttt{REP}_n = (1-\frac{\#\texttt{unique n-grams}(\hat{\textbf{x}})}{\#\texttt{total n-grams}(\hat{\textbf{x}})})$. $\hat{\textbf{x}}$ is the generated continuation. A higher diversity score indicates that generated outputs contain fewer repetitive segments. 

    \item \textbf{MAUVE}~\cite{pillutla2021mauve} measures the distribution similarity between the generated texts and reference texts. 

    \item \textbf{Coherence}~\cite{su2022a} is defined as the cosine similarity between the embeddings of the prompt $\mathbf{x}$ and the generated continuation $\hat{\mathbf{x}}$: $\texttt{COH}=\frac{f(\mathbf{x})f(\hat{\mathbf{x}})}{\left \| f(\mathbf{x}) \right\| \left \| f(\hat{\mathbf{x}}) \right\|}$, where $f$ is the SimCSE~\cite{gao-etal-2021-simcse} sentence embedding function.
\end{itemize}

For \textbf{human evaluation}, we conduct blind A/B tests with the help of proficient English speakers from a third-party grading platform. In the process of annotation, annotators are asked to compare two continuations of the same prompt and decide which one is better (or two are equally good/bad) by jointly considering fluency, coherence, and commonsense. Each case is rated by three annotators and we use majority vote.

\paragraph{Implementation Details}
For clarity, the variant of FSD using the vectorized $n$-gram model is named as FSD-vec. We set $n$ to 3 and 2 for FSD and FSD-vec respectively and $k$ to 6 for both variants. Based on our preliminary experiments, the penalty strength $\alpha$ is set to 3 and 1 for FSD and FSD-vec respectively. We find this setting is quite robust and generalizes well to different scenarios. 
\paragraph{Baselines}
To show the superior performance of FSD/FSD-vec, we mainly compared it with two recent search-based decoding methods, CD \cite{li2022contrastive} and CS \cite{su2022a}, since they were reported to outperform other existing decoding methods.\footnote{We omit greedy and beam search due to space limitation. The text generated by these two methods is very repetitive and of low quality \cite{li2022contrastive}.} We follow the suggested hyper-parameter settings from their respective papers. We also compare with top-$p$ sampling \cite{Holtzman2020The} because it is the most popular decoding method for open-ended text generation. We also include the results of typical sampling~\cite{meister2022typical}. We set $p$ in top-$p$ sampling and typical sampling to $0.95$, as adopted by \citet{li2022contrastive}.

\subsection{Main Results}
\label{sec:results}
\paragraph{Automatic Evaluation Results}
For automatic metrics, we believe that results closer to human are better because a higher score does not always indicate a better generation. For example, a random token sequence would obtain an extremely high diversity score, and a continuation identical to the input prompt would get a full coherence score. This is also commonly adopted in previous works \cite{Holtzman2020The,meister2022typical,xu2022learning}. 
Therefore, we highlight the results that are closest to human in our experiments.
From Table~\ref{tab:auto}, we can observe that:
\begin{itemize}[wide=0.\parindent,noitemsep,topsep=0.em]
    \item For diversity (\textbf{div}), FSD/FSD-vec matches or outperforms all other decoding baselines in six/five out of nine settings (the combinations of three LMs and three domains). In cases where FSD and FSD-vec are not the best, the gaps between them and the best scores are minimal ($<$ 0.03). It is worth noting that recent state-of-the-art methods (CD and CS) are very sensitive to the choices of the LMs. For example, CS fails to achieve reasonable diversity scores on all three benchmarks when using GPT2-Medium. The reason is that CS relies on the isotropy of the LM's latent space and GPT2-Medium may not fulfill this requirement. The diversity scores of CD also decrease significantly as the LM switches from GPT2-XL to GPT2-Medium, perhaps because the difference between the LM and its amateur is not sufficiently indicative of degeneration. In contrast, FSD and FSD-vec are much more stable in diversity. We attribute the success to that the operations of the anti-LM in FSD are relatively independent to the choice of the LM.

    \item For coherence (\textbf{coh}), FSD/FSD-vec achieves the best coherence scores in seven/four out of nine settings. These results emphasize the effectiveness of FSD and FSD-vec in generating coherent and contextually-appropriate continuations. We can see that sampling-based methods (top-$p$ and typical sampling) often deliver lower coherence scores than search-based methods (CS and CD). This confirms that sampling-based methods produce lexically diverse text at the expense of topic drift. Importantly, FSD and FSD-vec often attain better diversity and better coherence at the same time, suggesting our methods provide a better trade-off between diversity and coherence.

    \item For MAUVE (\textbf{mau}), sampling-based methods (particularly top-$p$ sampling) are generally better than search-based methods (CS, CD, FSD, and FSD-vec) though the gaps are often very small. However, it has been reported that the generation quality of CS and CD is better according to human evaluation. This indicates that MAUVE may not be a reliable metric which is also pointed out by \citet{suyixuan_emperical}. Therefore, we turn to extensive human evaluation.
\end{itemize}
\paragraph{Human Evaluation Results}
For human evaluation, we randomly select 100 prompts from each of the three benchmarks. We first compare FSD against top-$p$ sampling 
and two recent state-of-the-art methods, CS and CD. The results are shown in Table~\ref{tab:human_1}. We can see that on average across settings, annotators prefer FSD 1.30x more than CD, 1.26x more than top-$p$ sampling and 1.14x more than CS. FSD wins all the comparisons with the only exception: FSD vs CS on book. The results show that CS is the most competitive method, we then turn to compare FSD-vec with CS and FSD. As shown in Table~\ref{tab:human_2}, FSD-vec wins all the comparisons against CS and is preferred 1.49x more than CS. The quality of FSD-vec is on par with FSD.

\paragraph{Case Study}
We find that, compared with CS, FSD is less likely to generate continuations that deviate from the topic. Table~\ref{tab:case} shows two continuations from CS and FSD respectively. The prefix's topic is ``a musician is considering running for presidency''. But the topic of CS's output is concert tours which is irrelevant to that of the prefix. It may be because CS tends to excessively penalize tokens in comparison to FSD. For instance, CS has the potential to penalize tokens that have never occurred in the preceding context, as long as they produce similar hidden states. In contrast, FSD only penalizes tokens that appear in the context and genuinely result in repetitions.

\paragraph{Effect of Decoding Length}

\begin{figure}[t]
     \centering
     \includegraphics[width=0.8\linewidth]{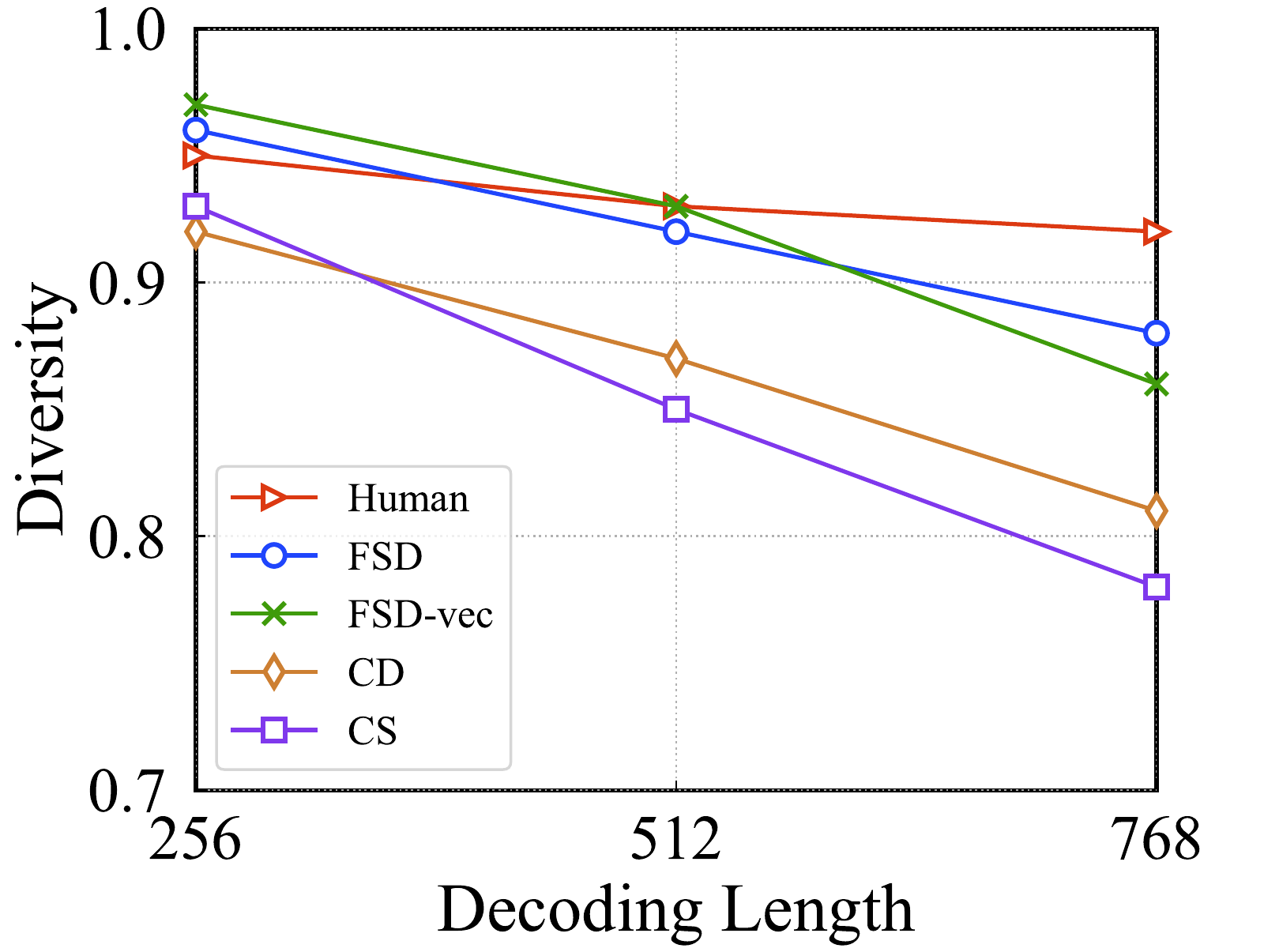}
        \caption{Diversity across different generation lengths.}
        \label{fig:len_effect}
\end{figure}

Next, we investigate the robustness of our methods in addressing the degeneration issue under different generation lengths. In Figure~\ref{fig:len_effect}, we present the diversity scores of FSD, FSD-vec, CS and CD when the generation length is 256, 512 and 768. As seen, the diversity of human-generated text is most stable across different lengths. The diversity of CS and CD drops dramatically as the generation length increases, resulting in a progressively larger disparity between the generated text and human-generated text. In contrast, FSD has the smallest slope and FSD-vec exhibits a similar slope to FSD from 256 to 512, and slightly steeper from 512 to 768. It reveals that our method is much more robust in reducing repetitions in longer sequence generation.

\begin{table}[t]
    \small
	\centering
	\scalebox{1.0}{
	\begin{tabular}{cccccc}
	 \toprule    
     \multirow{4}{*}{\rotatebox[origin=c]{90}{{\textbf{wikinews}}}}&\multicolumn{2}{c}{A is better}&Neutral&\multicolumn{2}{c}{B is better}\\
	    \cmidrule(lr){2-3}
	    \cmidrule(lr){4-4}
	    \cmidrule(lr){5-6}
            &FSD & \textbf{41\%} & 22\%  &37\%&top-$p$\\ 
        &FSD & \textbf{45\%}$^\dag$ & 25\%  &30\%&CS\\ 
        &FSD &\textbf{52\%}$^\dag$  & 12\% &36\% &CD\\
     \midrule 
	    \multirow{4}{*}{\rotatebox[origin=c]{90}{{\textbf{wikiext}}}}&\multicolumn{2}{c}{A is better}&Neutral&\multicolumn{2}{c}{B is better}\\
	    \cmidrule(lr){2-3}
	    \cmidrule(lr){4-4}
	    \cmidrule(lr){5-6}
           &FSD & \textbf{46\%}$^\dag$ & 24\%  &30\%&top-$p$\\ 
        &FSD & \textbf{39\%} & 24\% &37\% &CS\\ 
        &FSD&\textbf{37\%} &30\% &33\% &CD\\
        \midrule
	    \multirow{4}{*}{\rotatebox[origin=c]{90}{{\textbf{book}}}}&\multicolumn{2}{c}{A is better}&Neutral&\multicolumn{2}{c}{B is better}\\
	    \cmidrule(lr){2-3}
	    \cmidrule(lr){4-4}
	    \cmidrule(lr){5-6}
            &FSD &\textbf{41\%} & 24\% & 35\%&top-$p$\\ 
        &FSD &38\% &22\%  & \textbf{40\%}&CS\\ 
        &FSD &\textbf{46\%}$^\dag$ &19\% & 35\%&CD\\
     \bottomrule
	\end{tabular}}
    \caption{Human evaluation results of FSD. $^\dag$ means the advantage is statistically significant as judged by Sign Test with $p$-value$<0.05$.}
	\label{tab:human_1}
\vspace{-1em}
\end{table}
\begin{table}[t]
    \small
	\centering
	\scalebox{1.0}{
	\begin{tabular}{cccccc}
	 \toprule    
     \multirow{3}{*}{\rotatebox[origin=c]{90}{{\textbf{wikinews}}}}&\multicolumn{2}{c}{A is better}&Neutral&\multicolumn{2}{c}{B is better}\\
	    \cmidrule(lr){2-3}
	    \cmidrule(lr){4-4}
	    \cmidrule(lr){5-6}
          &FSD-vec &\textbf{44\%}$^\dag$ &25\% &31\% &CS\\
        &FSD-vec &36\%  & 21\% & \textbf{43\%}&   FSD\\
     \midrule 
	    \multirow{3}{*}{\rotatebox[origin=c]{90}{{\textbf{wikiext}}}}&\multicolumn{2}{c}{A is better}&Neutral&\multicolumn{2}{c}{B is better}\\
	    \cmidrule(lr){2-3}
	    \cmidrule(lr){4-4}
	    \cmidrule(lr){5-6}
          &FSD-vec &\textbf{51\%}$^\dag$ &26\%&23\% &CS\\
        &FSD-vec &\textbf{36\%}$^\dag$ & 33\%&31\% &FSD\\
        \midrule 

	    \multirow{3}{*}{\rotatebox[origin=c]{90}{{\textbf{book}}}}&\multicolumn{2}{c}{A is better}&Neutral&\multicolumn{2}{c}{B is better}\\
	    \cmidrule(lr){2-3}
	    \cmidrule(lr){4-4}
	    \cmidrule(lr){5-6}
          &FSD-vec &\textbf{42\%} &20\% &38\% &CS\\
        &FSD-vec &\textbf{37\%} & 27\% & 36\% &  FSD\\

     \bottomrule
	\end{tabular}}
    \caption{Human evaluation results of FSD-vec. $^\dag$ means the advantage is statistically significant as judged by Sign Test with $p$-value$<0.05$.}
	\label{tab:human_2}
 \vspace{-1em}
\end{table}

\begin{table*}[ht]
\begin{tabularx}{\textwidth}{X}
\toprule
\small \textbf{Prompt:} \footnotesize Following months of rumours, musician Wyclef Jean today confirmed that he is considering standing for the presidency of Haiti in the country's November elections. A statement from \\
\midrule 
\small \textbf{CS:} 
\footnotesize Jean's publicist reads: "Wyclef will make a decision on his political future in the coming days and is consulting with his family and advisors. "The announcement comes after Jean, who has been touring the world in support of his latest album A New Day In The Old Town, was forced to cancel a concert in the Dominican Republic last week due to security concerns. \textcolor{red}{He had been due to play the concert in Santo Domingo on November 4 but the concert was called off after Dominican authorities raised concerns about the safety of the concert venue, which is close to the border with Haiti. $\cdots$} \\

\midrule  
\small \textbf{FSD:} 
\footnotesize Jean's management company, La Nuit Productions, said: "Wyclef is currently in the process of gathering information and will make a decision in the coming days."Jean, who has been a vocal supporter of the Haitian government, has previously said he would consider running for the position. The musician, who is also a member of the hip-hop group A Tribe Called Quest, has not yet announced his candidacy. Haiti's current president, Michel Martelly, is widely expected to run for re-election. $\cdots$ \\

\bottomrule
\end{tabularx}
\caption{Case study: FSD vs CS.} 
\label{tab:case}
\end{table*}


\subsection{Decoding Speed}
\label{sub_sec:speed}
\begin{figure}[t]
    \centering
    \includegraphics[width=0.8\linewidth]{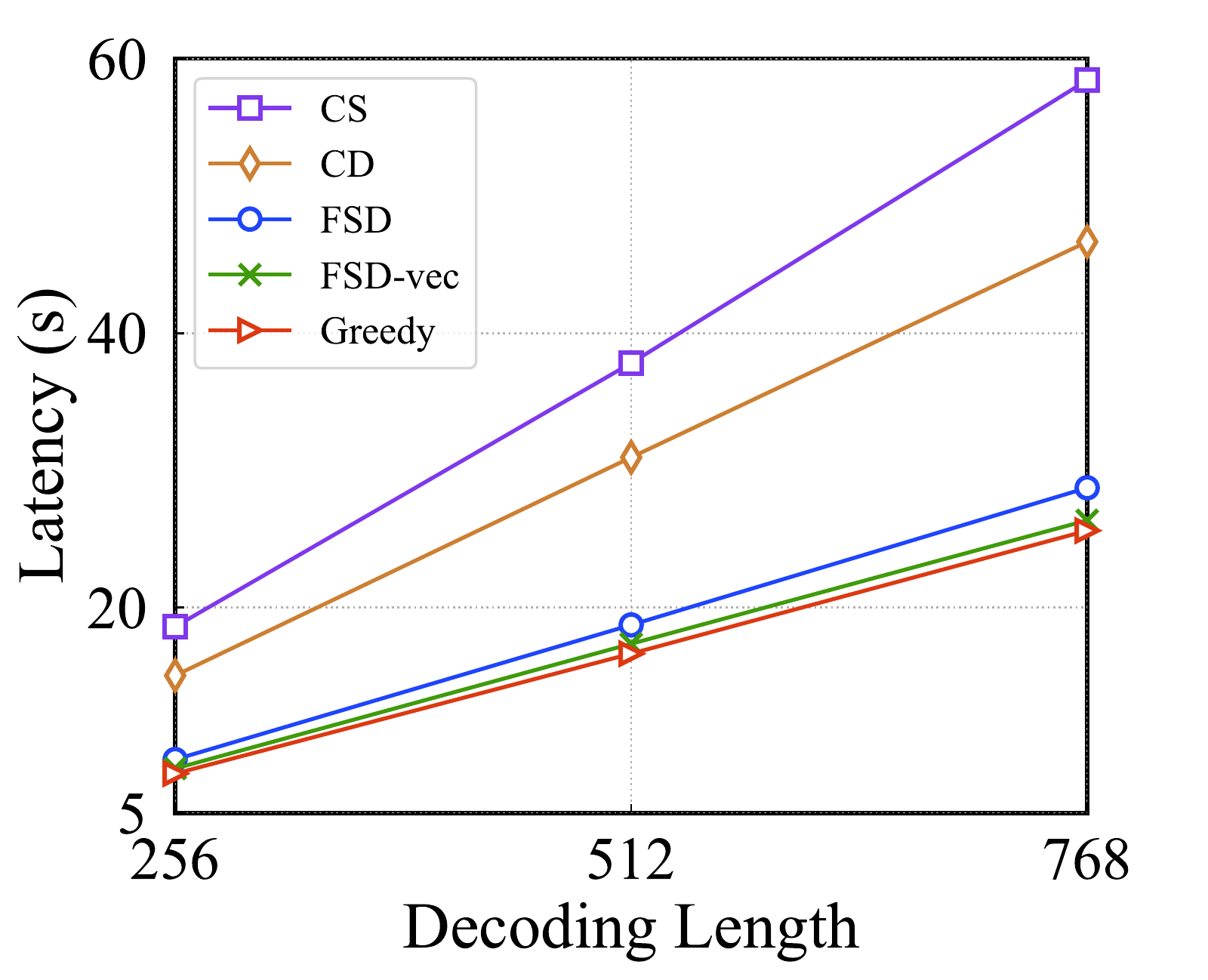}
    \caption{Decoding latency tested on GPT2-XL.}
    \label{fig:speed}
\end{figure}

To compare the decoding speed of different methods, we plot the decoding latency (seconds per instance) of search-based methods in Figure~\ref{fig:speed}. For clarity, we omit the results of sampling-based methods because they are close to greedy search. We can see that both FSD and FSD-vec demonstrate superior decoding speed compared with CD and CS, being more than 1.5x faster. In fact, FSD and FSD-vec can match the speed of greedy search. This can be attributed to the minimal computational overhead brought by the $n$-gram anti-LM, as opposed to the time-consuming look-ahead mechanism in CS and the running of an amateur LM in CD. Importantly, as the generation length increases, the absolute speed gap between FSD and CS/CD becomes even more pronounced, increasing from 8/10 seconds to 20/40 seconds per instance. This highlights the great efficiency advantage of our methods in generating long sequences. Note that FSD-vec is slightly faster than FSD. The reason is that the computation of the vectorized $n$-gram can be efficiently performed on GPUs.

 \begin{table*}[th]
   \small
    \centering
    
    \begin{tabular}{cc|ccc|ccc|ccc|ccc}
    \toprule
    & &\multicolumn{3}{c}{german}&\multicolumn{3}{c}{french}&\multicolumn{3}{c}{japanese}&\multicolumn{3}{c}{chinese} \\
        & &\textbf{div} &\textbf{mau} &\textbf{coh}  &\textbf{div} &\textbf{mau} &\textbf{coh}  &\textbf{div} &\textbf{mau} &\textbf{coh} &\textbf{div} &\textbf{mau} &\textbf{coh}\\
       \midrule  
       & Human &0.93 &1 &0.77  &0.88 &1 &0.65 & 0.56 &1 &0.60 &0.75 &1 &0.87 \\

      \midrule 
    \multirow{3}{*}{\begin{sideways} \scriptsize BLOOM\end{sideways}}
  
    &p=0.95   &0.93 &0.45 &0.69 &0.90 &\textbf{0.95} &0.63 &\textbf{0.56} &0.08 &0.51 &0.72 &\textbf{0.82} &\textbf{0.86}  \\
    &CS  &   0.20& 0.72& 0.72&  0.50& 0.90& 0.64& 0.04& \textbf{0.23}& 0.53 &0.07& 0.57& 0.80\\
    &FSD & \textbf{0.93} &\textbf{0.79} &
       \textbf{0.76}&\textbf{0.90} &0.90 &0.67 &0.59 &0.11 &\textbf{0.53}&\textbf{0.75} &0.75 &0.84 \\
    &FSD-vec & 0.92 &0.73 &0.74 &0.91 &0.91 &\textbf{0.65} &0.55 &0.06 &0.50&0.80 &0.75 &0.83 \\  
       \midrule 
    \multirow{3}{*}{\begin{sideways} \scriptsize OPT\end{sideways}}
    &p=0.95 &0.91 &\textbf{0.70} & 0.73&\textbf{0.89} &0.70 &0.60&0.73 &0.61 &0.55&-&-&-\\
       &CS&0.83 &0.60 &0.72&0.84 &0.72 &0.60 &0.42&0.18&0.53&-&-&-\\
       &FSD&\textbf{0.93} &0.69 &
       \textbf{0.73}&0.91 &\textbf{0.73} &\textbf{0.62}&0.65&\textbf{0.69}&\textbf{0.59}&-&-&-\\
       &FSD-vec&0.93 &0.64 &0.73&0.85 &0.69 &0.61&\textbf{0.64}&0.58&0.56&-&-&-\\

    \midrule 
    \multirow{3}{*}{\begin{sideways} \scriptsize LLaMA\end{sideways}}
     &p=0.95 &0.94 &0.94 &0.75 &0.90 &\textbf{0.93} & 0.64&-&-&-&-&-&-\\
       &CS&0.90 &0.78 &0.73 & \textbf{0.90} &0.73 &0.61&-&-&-&-&-&-\\
       &FSD&\textbf{0.93} &0.88 &
       \textbf{0.75} &0.93 &0.85 &\textbf{0.65}&-&-&-&-&-&-\\
       &FSD-vec&0.92 &\textbf{0.94} &0.74 &0.91 &0.88 &0.64&-&-&-&-&-&-\\
   
    \bottomrule
    \end{tabular}
    
    \caption{Automatic evaluation results on four non-English datasets and three LMs.}
     \label{tab:auto_ol}
\end{table*}

 \begin{table}[th]
   \small
    \centering

    \resizebox{0.9\linewidth}{!}{\begin{tabular}{c|ccc|ccc}
    \toprule
    & \multicolumn{3}{c}{BLOOM}&\multicolumn{3}{c}{OPT}\\
        & \textbf{R-1} &\textbf{R-2} &\textbf{R-L}  &\textbf{R-1} &\textbf{R-2} &\textbf{R-L}  \\
        \midrule 
        beam (size=8)&32.0 &6.7 &27.8& 35.8 &5.9 &31.2\\ 
        p=0.95 &27.5 &2.9 &23.6 &24.1 &2.5 &20.7\\
        CS   & 34.1 &5.5 &30.4 &35.6 &8.3 &31.3\\
        FSD &\textbf{34.2} &\textbf{5.9} &\textbf{31.3}& \textbf{37.4} &\textbf{9.8} & \textbf{33.7}\\
        FSD-vec &33.2 & 5.2 &29.1& 37.1 &8.6 &32.1\\ 
	    
       
    \bottomrule
    \end{tabular}}
    
    \caption{Automatic evaluation results on XSum.}
     \label{tab:auto_sum}
\end{table}

\subsection{Universality}
\label{sec:uni}
\paragraph{More Languages, Models and Datasets} So far, our evaluation has been primarily focused on English corpora, and the types of LMs used are also limited. We here expand our evaluation to include other non-English languages using various LMs.  
We conduct experiments on four datasets, chinese-wiki~\footnote{\url{https://github.com/SigmaQuan/Awesome-Chinese-Corpus-Datasets-and-Models}}, japanese-news~\footnote{\url{https://www.kaggle.com/datasets/tanreinama/japanese-fakenews-dataset}}, german-wiki, and french-wiki~\footnote{\url{https://huggingface.co/datasets/wikipedia}}. 
We adopt a variety of popular LMs, including BLOOM-7b~\cite{workshop2023bloom}, LLaMA-7b~\cite{touvron2023llama}, OPT-6.7b~\cite{zhang2022opt}. 
The evaluation results are shown in Table~\ref{tab:auto_ol}, where also report the results of the state-of-the-art
decoding methods, CS and top-$p$ (the missing positions indicate the LM does not support the language.).  As seen, FSD and FSD-vec generally outperform CS and top-$p$ (most of the boldfaced numbers are from FSD and FSD-vec.).
It should be noted that for BLOOM-7b, CS does not work entirely in all four languages (see the extremely low diversity scores).  Additionally, the performance of CS also exhibits greater sensitivity to different languages. For instance, when applied to the japanese dataset using OPT, the diversity and MAUVE scores are notably low. In contrast, FSD and FSD-vec deliver much more stable performance across different settings, indicating FSD/FSD-vec can be a universal choice for open-ended text generation.

\paragraph{Instruction Following}
The latest generation of LLMs such as ChatGPT \cite{2022OpenAIchatgpt} and LLaMA-2-chat~\cite{touvron2023llama2} have the capabilities to perform various tasks by following natural language instructions. This instruction-following approach has become the standard form for harnessing LLMs. Therefore, we compare FSD/FSD-vec against baselines within this context. Specifically,  we follow the widely accepted comparison setting ~\cite{alpaca_eval}, i.e., reporting the win rates against text-davinci-003 on the alpaca\_eval dataset\footnote{\url{https://huggingface.co/datasets/tatsu-lab/alpaca_eval}} with the help of GPT-4 \cite{2023GPT4Openai}. We adopt the LLaMA-2-7b-chat model~\cite{touvron2023llama2} since it is among the most popular instruction-tuned models. The results of different decoding methods are shown in Table~\ref{tab:gpt4}. The results clearly indicate that FSD/FSD-vec outperforms the baselines, thus further validating the effectiveness of our approach.


 \begin{table}[t]
   \small
    \centering

    \resizebox{0.9\linewidth}{!}{\begin{tabular}{cccccc}
    \toprule
       \textbf{method}&top-p&CD&CS&FSD&FSD-vec\\
       \midrule 
       \textbf{win rate}&77.20&-&78.32&82.32&81.84\\
    \bottomrule
    \end{tabular}}
    
    \caption{Win rate of GPT-4 evaluation. CD is omitted since it requires a smaller amateur model and the model we use is already the smallest one.}
     \label{tab:gpt4}
    \vspace{-1em}
\end{table}

\paragraph{Close-Ended Generation Task: Summarization}
So far, our evaluation has been focused on open-ended text generation and general-purpose instruction following. We also evaluate our methods on a specific, close-ended generation task: summarization. We use the XSum dataset~\cite{Narayan2018DontGM}. As shown in Table~\ref{tab:auto_sum}, FSD/FSD-vec is generally better than other baselines. 
It showcases that our methods can also work well in close-ended scenarios.

\subsection{Hyperparameter Analysis}
\paragraph{Analysis of $\alpha$}
We first study the effect of the penalty strength $\alpha$. We present the results in Table~\ref{tab:alpha}.  We notice that as $\alpha$ increases, the div score consistently increases. This is an expected outcome, as a larger $\alpha$ imposes a greater penalty on repetitive content, thereby promoting increased diversity in the model's outputs. The coh score demonstrates a decreasing trend. The reason is that penalizing the most probable tokens may damage the coherence between the prefix and the continuation. Consequently, we see that the mauve score initially shows an upward trend and then experiences a slight decrease.

\paragraph{Analysis of $n$}
We study the effect of the hyperparameter $n$ as shown in Table~\ref{tab:n}. We can observe that diversity and coherence are very stable for different $n$, when $n>3$, the mauve begins to decrease.

\paragraph{Analysis of $k$}
We investigate the impact of the hyperparameter $k$, as presented in Table~\ref{tab:k}. When $k$ is assigned a minimal value, a notably lower diversity (div) is observed. This can be attributed to the reduced search space associated with a smaller $k$, which consequently constrains the diversity of generated outcomes. Conversely, upon incrementing $k$ past a specific threshold, all evaluated metrics—diversity, mauve, and coherence—demonstrate substantial stability, with only negligible fluctuations observed. This stability suggests that the effective selection space of FSD predominantly comprises a limited number of top tokens.

Despite that the hyperparameters can take different values, we recommend using the default settings of those hyperparameters and only adjusting $\alpha$ to suit different tasks.

 \begin{table}[t]
   \small
    \centering

    \begin{tabular}{c|ccc}
    \toprule
       &\textbf{div}&\textbf{mau}&\textbf{coh}\\
       \midrule 
       $\alpha=1$& 	0.62& 	0.88& 	0.67\\
       $\alpha=2$&     0.87& 	0.94& 	0.66\\
       $\alpha=3$&     0.93& 	0.93& 	0.66\\
       $\alpha=4$&     0.95& 	0.89& 	0.65\\
    \bottomrule
    \end{tabular}
    
    \caption{Analysis of $\alpha$. The experiments are conducted on wikinews using GPT2-XL with FSD.}
     \label{tab:alpha}
\end{table}

\begin{table}[h!]
\centering
\begin{tabular}{c|ccc}
\hline
& \textbf{div} & \textbf{mauve} & \textbf{coh} \\
\hline
$n=2$ & 0.93 & 0.93 & 0.64 \\
$n=3$ & 0.93 & 0.94 & 0.65 \\
$n=4$ & 0.92 & 0.88 & 0.65 \\
\hline
\end{tabular}
\caption{Analysis of $n$. The experiments are conducted on wikinews using GPT2-XL with FSD.}
\label{tab:n}
\end{table}

\begin{table}[h!]
\centering
\begin{tabular}{c|ccc}
\hline
& \textbf{div} & \textbf{mauve} & \textbf{coh} \\
\hline
$k=2$ & 0.69 & 0.90 & 0.66 \\
$k=4$ & 0.91 & 0.92 & 0.65 \\
$k=6$ & 0.93 & 0.93 & 0.64 \\
$k=8$ & 0.94 & 0.94 & 0.64 \\
$k=10$ & 0.94 & 0.94 & 0.64 \\
\hline
\end{tabular}
\caption{Analysis of $k$. The experiments are conducted on wikinews using GPT2-XL with FSD.}
\label{tab:k}
\end{table}

\section{Conclusion and Future Directions}
We proposed FSD, an effective, efficient, and universal decoding method for avoiding the degeneration problem and improving generation quality. FSD constructs an anti-LM on-the-fly to penalize repetitive generation. Extensive evaluations and analyses confirm its effectiveness across open-ended text generation, instruction following, and summarization tasks. In addition, FSD demonstrates better efficiency and generality compared with existing state-of-the-art decoding methods.

An intriguing future research direction could involve a more nuanced approach to repetitions. In fact, some  grams (like named entities) might not require penalization at all. Therefore, researchers may develop more meticulous algorithms based on FSD to discern the contexts and conditions under which repetitions should be penalized. This would enable a more refined and context-sensitive application of repetition management in text generation.

\section*{Ethics Statement}

Due to the nature of language models, we note that the generations of our method may have offensive, toxic, unfair, or unethical content. The generations of our method may also have hallucinated content and can be misleading. When deployed in real-world applications, special attention should be paid to avoid inappropriate generations. For example, one can use post-process steps such as toxicity identification and fact checking.

\section{References}
\label{sec:reference}
\bibliographystyle{lrec-coling2024-natbib}
\bibliography{lrec-coling2024-example}

\newpage
\clearpage
\appendix
\section{Implementation Details}
\label{sec:app_parasetting}
We provide the detailed pseudo code of FSD in Alg. \ref{alg:fsd}.

\paragraph{Stopwords and Punctuations}Stopwords significantly influence the diversity of sentence structures as they often appear at the beginning of sentences, such as "The..." or "He...". To provide finer control over the penalty applied to stopwords, we introduce a discount factor $\phi$. This factor is multiplied by the second term of Eq.~\ref{eq:fsd}, replacing $\alpha$ with $\phi \cdot \alpha$ specifically for stopwords.
A smaller $\phi$ tends to produce sentences with similar structures, as demonstrated in the example provided in Table~\ref{tab:case2}. Conversely, a larger $\phi$ can lead to the generation of invalid sentences due to the heavy penalty imposed on stopwords at the beginning of a sentence. This may result in the selection of incorrect tokens, as illustrated in the example presented in Table~\ref{tab:case_sw_1}. 

We also experimentally find that penalizing punctuations can sometimes introduce grammar errors in the generated text.  Specifically, when utilizing GPT2 as the base model, we have found that the punctuation symbols \texttt{ĊĊ} (representing "\textbackslash n \textbackslash n") and \texttt{Ċ} (representing "\textbackslash n") have a significant impact on the grammatical correctness of the output. An example illustrating this phenomenon is provided in Table~\ref{tab:case_punt}. 
In our experiments, we do not penalize punctuations and the punctuations set is $\mathcal{P}=\left \{ \texttt{. , : " ` ĊĊ Ċ} \right \}$.


\begin{algorithm}[t]
\small
   \SetAlCapFnt{\small}
   \SetAlCapNameFnt{\small}
    \caption{FSD Decoding}
    \SetKwInOut{Input}{Input}
    \SetKwInOut{Output}{Output}
    \label{alg:fsd}
    
    \Input{the LM $ p_\theta$ (e.g. GPT2); the anti-LM $ p_\omega$; the prompt text $x_{1:l}$; the decoding length $m$; the stopwords set $\mathcal{S}$; the punctuation set $\mathcal{P}$;}
    Construct the anti-LM $ p_\omega$  with the prompt $x_{1:l}$;\\
    \For{step $t=l+1$ \textbf{to} $l+m$}
    {
        Compute next token distribution $p_{\theta}(\cdot|x_{< t})$;\\
       
        Get $V^{(k)}$ from $p_{\theta}(\cdot|x_{< t})$;\\
        \For { candidate $v\in V^{(k)}$} 
        {
        Get the penalty $p_\omega(v|x_{< t})$ according to Eq.~\ref{eq:interpolation} (discrete version) or Eq.~\ref{eq:vectorized} (vectorized version); \\
       
        }
    $x_t = \argmax_{v\in V^{(k)}}\{ \mathrm{FSD}(v|x_{<t})\}$;\\
    Update $p_\omega$ with $x_t$;\\ 
    }
    \Output{The generated text $\hat{x}$.}
\end{algorithm}
\paragraph{Hyper-parameter Settings}
 We search $\phi$ from $\{0.2, 0.4, 0.6, 0.8, 1, 1.5\}$. The detailed parameter settings are listed in Table~\ref{tab:settings}.
\begin{table}[h]
    \small
    \centering
    \resizebox{0.6\linewidth}{!}{\begin{tabular}{c|ccc}
      \toprule
       &$n$ &$\alpha$ &$\phi$ \\
         \midrule
      &\multicolumn{3}{c}{FSD}\\
      \cline{2-4}  
       wikinws&  3 &3   &0.2   \\
       wikitext& 3 &3   &0.4   \\
       book&     3 &3   &0.6   \\
       \midrule
        &\multicolumn{3}{c}{FSD-vec}\\
        \cline{2-4}
        wikinws&  2 &1  &0.2   \\
       wikitext& 2 &1  &0.2   \\
       book&     2 &1  &0.6   \\
      \bottomrule
    \end{tabular}
    }
    \caption{Parameter settings of $\phi_s$} 
    \label{tab:settings}
\end{table}


\section{Further Analysis}
\label{sec:app_furnan}

\subsection{Effect of Smoothing}
\label{sec:app_es}
\begin{figure}[h]
\setlength{\abovecaptionskip}{1pt}
    \centering
    \includegraphics[width=0.78\linewidth]{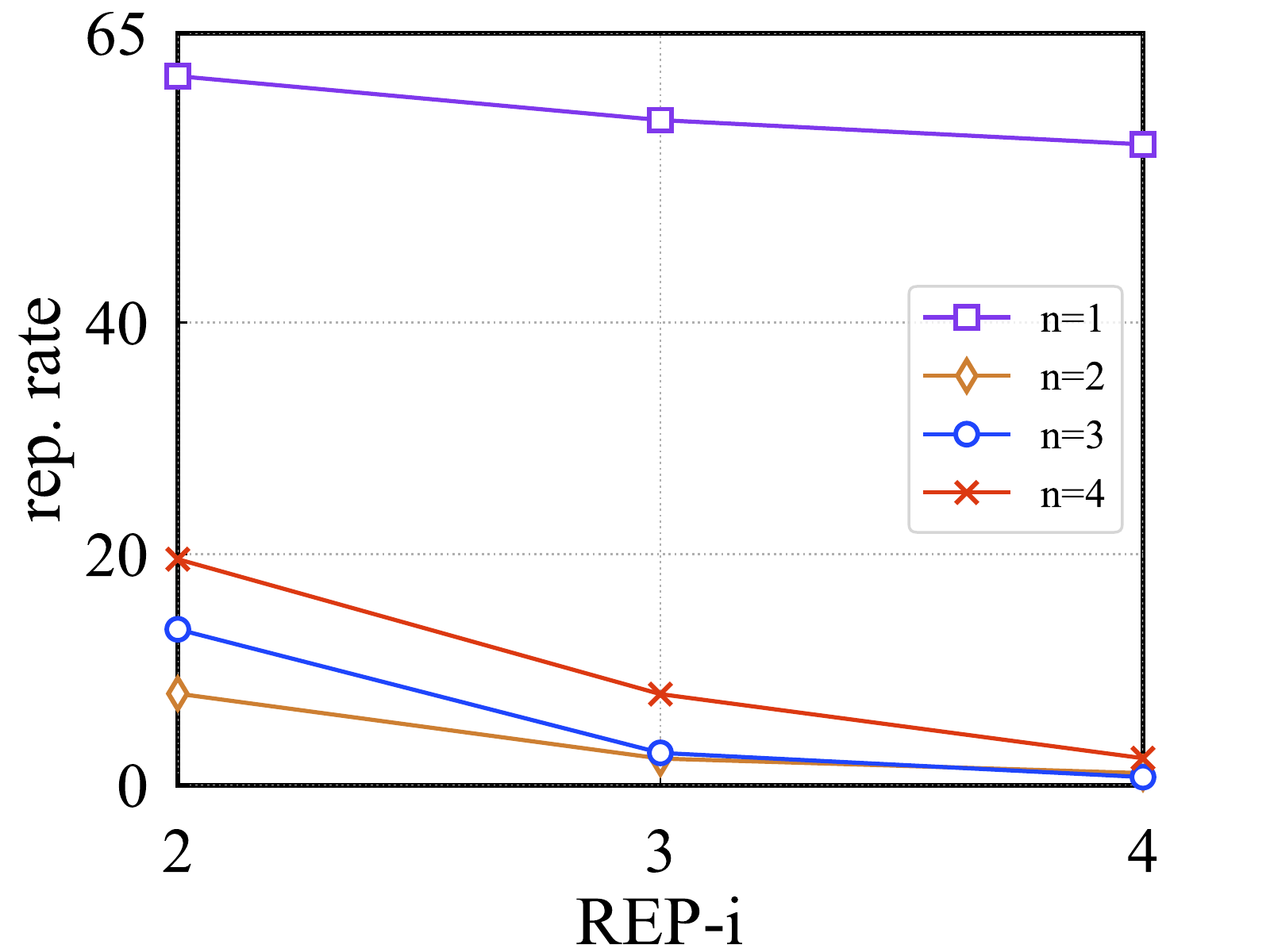}
    \caption{Repetition rate for different $n$.}
    \label{fig:normal_ngram}
    \vspace{-1.5em}
\end{figure}

In previous experiments, we adopt the smoothed $n$-gram model as the anti-LM.
To understand the effect of smoothing, we also implement the unsmoothed $n$-gram and run multiple experiments with different $n\in [1, 2, 3, 4]$. Then, we calculate $\texttt{REP-i},i \in [2,3,4]$. The results are illustrated in Figure~\ref{fig:normal_ngram}. We found that if unsmoothed $n$-gram is applied, the best performance is achieved when $n=2$. The reason for this phenomenon is that if $n>2$, the unsmoothed $n$-gram LM can not penalize grams with lengths smaller than $n$, which is manifested by the high $\texttt{REP-i}, i<n$ in Figure~\ref{fig:normal_ngram}.  However, setting $n=2$ sometimes may not be a good option due to the BPE encoding algorithm, under which a word (e.g., name of a person) can be decomposed into multiple tokens. If penalized heavily, these words may not be recovered. 

\section{More Cases}
\label{sec:app_morecase}
\subsection{FSD vs CS}

We provide more cases on Table~\ref{tab:case_wikinews}, Table~\ref{tab:case_wikitext}, Table~\ref{tab:case_book} for wikinews, wikitext and book respectively to compare FSD with CS.

\subsection{FSD vs FSD-vec}
We provide more cases on Table~\ref{tab:case_wikinews_ours}, Table~\ref{tab:case_wikitext_ours}, Table~\ref{tab:case_book_ours} for wikinews, wikitext and book respectively to compare FSD with FSD-vec.

\begin{table*}[ht]
\setlength{\abovecaptionskip}{2pt}
\begin{tabularx}{\textwidth}{X}
\toprule
\small \textbf{Prompt:} \footnotesize Following months of rumours, musician Wyclef Jean today confirmed that he is considering standing for the presidency of Haiti in the country's November elections. A statement from \\
\midrule 
\small \textbf{CS:} 
\footnotesize Jean's publicist reads: "Wyclef will make a decision on his political future in the coming days and is consulting with his family and advisors. "The announcement comes after Jean, who has been touring the world in support of his latest album A New Day In The Old Town, was forced to cancel a concert in the Dominican Republic last week due to security concerns. He had been due to play the concert in Santo Domingo on November 4 but the concert was called off after Dominican authorities raised concerns about the safety of the concert venue, which is close to the border with Haiti. In an interview with Rolling Stone, Jean said he was "saddened" by the cancellation, but that he had no choice but to cancel the concert. "I'm a human being and it's my job to do what I have to do," he said. "If there's a risk, I have to take it. But I can't go to the Dominican Republic with this threat hanging over my head."Jean's tour, which has taken him to South Africa, the United Arab Emirates (UAE, where he performed at the Abu Dhabi Formula One Grand Prix in October) and Brazil, is the first time he has performed in Haiti since \\
\midrule  
\small \textbf{FSD:} 
\footnotesize Jean's management company, La Nuit Productions, said: "Wyclef is currently in the process of gathering information and will make a decision in the coming days."Jean, who has been a vocal supporter of the Haitian government, has previously said he would consider running for the position. The musician, who is also a member of the hip-hop group A Tribe Called Quest, has not yet announced his candidacy. Haiti's current president, Michel Martelly, is widely expected to run for re-election. In a statement, Jean said: \"I am considering running for president of Haiti. I have been following the election process closely and I believe that Haiti needs a new leader who can bring about change and rebuild the country. "I have always been a supporter of President Martello and his government. I am also a fan of the music of A Tribe called Quest and have supported them in the past. "The Haitian people have suffered greatly under the current government and I feel that the time has come for a change. "Hearing the news of the earthquake in Haiti, I was deeply saddened and saddened to hear of the loss of life and the destruction of homes and businesses.\\
\bottomrule
\end{tabularx}
    \caption{One case of FSD versus CS. The prefix's topic is ``a musician is condsiering runing for presidency''. But the topic of CS's output is concert tours which is irrelevant to that of the prefix. By contrast, FSD always sticks to the topic of the prefix.}
    \label{tab:full_case}
    \vspace{-1em}
\end{table*}

\begin{table*}[ht]

\begin{tabularx}{\textwidth}{X}
\toprule
\small \textbf{Prompt:} \footnotesize A gas explosion in the Grand Riviera Princess resort in Playa del Carmen, Mexico earlier today killed seven people and left 20 people injured, two critically according to \\
\midrule 
\small \textbf{Generations:} 
\footnotesize local media. The explosion occurred at around 8:30am local time (10:00am GMT) at the hotel's pool area, which is located on the second floor of the resort.The hotel is a popular destination for tourists and is known for its large pool and spa. The cause of the explosion is still unknown, but local authorities have said that the cause is not related to the fire that broke out in the pool earlier in the day. \textcolor{red}{The resort's management has said that} the incident is not connected to the blaze that erupted in the same area earlier in the day.\textcolor{red}{The fire} was extinguished by the resort's staff and the guests were evacuated. \textcolor{red}{The victims} of the blast have been identified as: \textcolor{red}{A man in his 30s A woman in her 20s A young man in his 20s A middle-aged woman A 40-year-old man A 50-year old woman A 60-something woman A 70-some-year women A man in his 50s A woman in her 30s A 30-something man A 20-something girl A 15-some year}\\
\midrule 
\small \textbf{Prompt:} \footnotesize According to data from South Korea's national bank, the country's economy grew by 0.2\% in the last quarter of 2009. The rise is now\\
\midrule 
\small \textbf{Generations:} 
\footnotesize expected to be 0.3\% in the first quarter of 2010. \textcolor{red}{The government} has been trying to boost the economy by cutting taxes and increasing spending.\textcolor{red}{The central bank} has also been cutting interest rates to stimulate the economy.\textcolor{red}{The Bank} of Korea has cut its benchmark interest rate to 1.5\% from 2.5\%.\textcolor{red}{The rate} is still higher than the rate of inflation.\textcolor{red}{The country} has been struggling with a huge debt burden.The debt is estimated at about \$1.4 trillion.\textcolor{red}{The South Korean government} has been attempting to reduce the debt by selling off state-owned companies.\textcolor{red}{The government} has been selling off the Hyundai Motor Group, the nation's largest carmaker, and the Korea Development Bank.\textcolor{red}{The state} has also been selling off its stake in the Samsung Group.\textcolor{red}{The sale} of the two companies has been a success.\textcolor{red}{The Hyundai Group} has been sold for \$8.6 billion.\textcolor{red}{The Korea DevelopmentBank} has been sold for about half that amount.\textcolor{red}{The two sales} have been a big success.\textcolor{red}{The sales} have helped the government to lower its debt to GDP ratio.\textcolor{red}{The debt} to the GDP is now down to about 50\% of GDP.\textcolor{red}{The Korean economy} is expected to grow by 1.\\ 

\bottomrule
\end{tabularx}
    \caption{No penalty on stopwords may procude sentences with similar grammatical structure.}
    \label{tab:case2}

\end{table*}

\begin{table*}[ht]
\begin{tabularx}{\textwidth}{X}
\toprule
\small \textbf{Prompt:}Charlie, a Chimpanzee in a South African zoo who became addicted to cigarettes, has died at the age of 52, exceeding the average lifespan of a captive \\   \small \textbf{Generation:} \footnotesize animal.  The cause of death was not immediately known.  Charlie, who was born in captivity in 1989, was diagnosed with lung cancer in 2011.  He was given a lung transplant in 2013, but his health deteriorated again and he was put down.  His death comes just weeks after another chimpanzee, Harambe, was shot dead by zoo officials after a child fell into his enclosure.  \textcolor{red}{Image copyright Getty Images Image caption Charlie was one of the last chimps to be born at an American zoo}  In a statement, the zoo said: "Charlie's passing is a great loss to the animal kingdom.  "Charlie was a beloved member of our family and we will miss him dearly."  Chimpanzees are the closest living relatives of humans, and are often used in research.  They are also used for entertainment, as they are highly intelligent and can mimic human speech.  Charlie, whose name means "chimney sweep" in Swahili, was born on 19 April 1989 at San Diego Zoo.  He lived at the San Francisco Zoo for five years before moving to Cincinnati Zoo in July 2014.  Image caption Chimps are known for their intelligence and mimicry  The \\
\midrule 
\small \textbf{Prompt:}Police have said that an explosion in a hotel garage in San Pedro, Los Angeles, California, that sparked an evacuation of a large area was an act of vandalism \\  \small \textbf{Generation:} \footnotesize .  The incident occurred at around 2:30am local time (11:00 GMT) on Thursday.  A fire broke out in the garage, which is located near the hotel, and caused a huge explosion.  \textcolor{red}{Image copyright Getty Images Image caption The hotel was evacuated after the explosion}  "It was a very loud explosion," said one witness.  "I was sleeping and woke up to the sound of the explosion. I looked out the window and saw a lot of smoke and fire."  Another witness, who did not want to be named, said: "I heard a loud bang and then I saw the fire.  "The fire was so big it was like a forest fire. It was just like something out of a movie."  Media playback is unsupported on your device Media caption Footage shows the aftermath of the blast  Los Angeles Mayor Eric Garcetti said the incident was "an act against our city".  "We will not tolerate this type of violence," he said.  "This is an attack on our way of life, our economy, our infrastructure, and our future."  He said he had ordered the evacuation "out of an abundance of caution".  "Our city \\
\bottomrule 
\end{tabularx}
    \caption{Penalizing stopwords heavily may produce invalid generations. The invalid generation ``Image copyright ...'' may occur in some examples because the stopwords that have a larger probability than ``Image'' have a larger penalty. Therefore, ``Image'' is selected. }
    \label{tab:case_sw_1}

\end{table*}

\begin{table*}[ht]

\begin{tabularx}{\textwidth}{X}
\toprule
\small \textbf{Prompt:} Apple Inc. announced Friday that it will give owners of its new iPhone 4 a free case in response to mounting concerns over the device's antenna placement. The iPhone \\   \small \textbf{Generation:} \footnotesize maker said it is offering the free cases to customers who have already purchased the phone.  The company said the new cases will be available in stores and online starting Friday. \textcolor{red}{Apple said the cases will have a "unique design" that will help protect the iPhone's antennas from damage. , which is made by Corning Inc., will be sold in the U.S. and Canada for \$29.99.} The company also said that the case will be made in China, where the company has been making its products for years.  "We are committed to providing our customers with the best possible products and we are always looking for ways to improve our products and services," Apple spokeswoman Kristin Huguet said in a statement. "We know that our fans want to protect their iPhone and this case is designed to do just that."  Apple's decision to offer the free protective cases comes after a number of reports of people experiencing problems with their iPhones' antennas. Some people have said they have been unable to make calls or send text messages because their phones were unable to receive signals. Others have complained that their devices have lost reception or have been damaged by the antenna.  a report by The Wall Street Journal last week that said some iPhone owners had been complaining \\
\midrule 
\small \textbf{Prompt:}Wikinews reporter Iain Macdonald has performed an interview with Dr Isabella Margara, a London-based member of the Communist Party of Greece (K \\  \small \textbf{Generation:} \footnotesize KE), who has been a member of the party since the 1970s.  Dr Margaras is a well-known figure in the Greek left, having been a prominent member of the KKE since the 1980s, and having served as a deputy to the party's leader, Dimitris Koutsoumbas. She is also a former member of the Socialist Party (PASOK) and a long-time member of the European Parliament. \textcolor{red}{In the interview, she discusses the current situation in Greece, the role of the EU in the country's crisis, and the possibility of a new left government in Athens.  and the role of the EU in the countrys crisis,.} In the article, Dr Margareta Margara also discusses the recent decision by the Greek government to impose capital controls on the banks, and the implications of this for the country. The interview was conducted by Iain MacDonald on behalf of The Independent.   Iain: What is the situation in your country?  Margaras: The situation is very bad. The government has imposed capital control on the banking system. This is a very serious situation. The banks are closed and there is no access to money. There is no money in the banks. There is \\
\bottomrule 
\end{tabularx}
    \caption{Penalizing punctuations can produce grammar errors.}
    \label{tab:case_punt}

\end{table*}

\begin{table*}[ht]
\begin{tabularx}{\textwidth}{X}
\toprule
\small \textbf{Prompt-1:} \footnotesize On Friday, the 189 Parties to the Nuclear Non-Proliferation Treaty (NPT) agreed to a final resolution calling for steps toward the banning of weapons \\ 
\midrule 
\small \textbf{FSD:} \footnotesize  of mass destruction.  The resolution, which was adopted by consensus, calls for the "immediate and complete cessation of all nuclear, chemical and biological weapons-related activities" and "the establishment of a comprehensive, verifiable and irreversible nuclear-weapons-free zone in the Middle East."  It also calls for "a comprehensive, effective and enforceable prohibition on the production, stockpiling, acquisition, transfer, use and threat of use of nuclear weapons and all other weapons of indiscriminate and disproportionate impact."  The NPT was created in 1968 to prevent the spread of nuclear and other arms.  "The international community has been working for many years to achieve a world without nuclear weapons," said U.S. Ambassador to the United Nations Samantha Power. "Today, we have taken a major step forward toward that goal."  "This is a historic day for the international non-proliferating movement," she said. "It is a day of hope and optimism for all those who believe that the world can be a safer, more peaceful place."  "We are proud of the work that has been done by the 193 N.G.O. member states to reach this historic agreement," Power said. "We will continue to \\
\midrule 
\small \textbf{CS:} \footnotesize of mass destruction.  In the meantime, the United States is continuing to develop nuclear weapons and ballistic missile defense systems that it can use against Iran if it were to violate the terms of the agreement.  On Monday, Secretary of State John Kerry said Iran's "threats and provocations are a clear violation of the spirit and letter" of the NPT, which was signed by 122 countries in the 1970s.  Kerry's comments came during a visit to South Korea, where he met with President Park Geun-hye and Foreign Minister Yun Byung-se.  "We have made it clear to the Iranians that their nuclear program must be subject to IAEA safeguards, and that they must cease work on a heavy water reactor that could produce plutonium," Kerry said.  The Obama administration has been working for more than a year to reach a deal with Iran that would curb its nuclear program in exchange for relief from economic sanctions.  But negotiations have hit snags, most recently over Iran's refusal to allow inspectors into Parchin, a military site where Iran is believed to have carried out experiments on nuclear weapons in the past.  Iranian officials have said the site is for research and development, but Western intelligence \\
\midrule
\midrule 
\small \textbf{Prompt-2:} \footnotesize Inventor Walter Frederick Morrison has died of the age of 90. He was famous for being the inventor of the flying disc, more commonly identified as the fr \\
\midrule 
\small \textbf{FSD:} \footnotesize isbee.  Morrison was born in New York City in 1891. His father was a lawyer and his mother was a homemaker. Morrison's father died when he was young, and he moved to Chicago to live with his aunt.  He attended the University of Chicago, where he studied engineering. In 1915, he joined the U.S. Army Air Corps, and was stationed in France during World War I.  In 1919, Morrison was commissioned as an officer in the Army Signal Corps. He served in the Signal Service, which was responsible for transmitting messages across the Atlantic.  After the war, Morrison went to work for the Bell Telephone Company, where his job was to design and build telephone systems.  In 1928, he founded the Morrison Electric Company, which manufactured and sold electric motors and generators.  In 1939, Morrison founded the American Aircraft Corporation, which made aircraft engines.  In 1941, the company was sold to the General Electric Corporation, and Morrison became president of the new company.  The company's first aircraft, the P-51 Mustang, was designed by Morrison.  The P2V Mustang, a fighter plane, was the first American fighter to be powered by a jet engine. \\
\midrule 
\small \textbf{CS:} \footnotesize sbee.  Morrison's name is on the back of every disc sold, and the discs are made in his hometown of St. Louis, Missouri.  The disc was invented in 1887 by a man named George Eastman, who patented it in the United States and sold it to Parker Brothers for \$1,500. It was a flimsy disc with a hole in the middle that could be thrown with one hand, according to Smithsonian Magazine.  In 1889, Parker Brothers began to sell the disc in Europe, and by the 1920s, there were over 2 million discs in circulation, according to Smithsonian Magazine.  But the disc was not without its problems. It was difficult to make, and many people complained about the disc breaking on the way to the basket.  "I have seen it, and it is terrible," one golfer said, according to Smithsonian Magazine. "The ball goes in, and you have to hold it in your hand for a long time, and then it breaks in the basket."  The frisbee, on the other hand, was much more durable and easy to throw.  "It's like a golf club in that way," Morrison told National Geographic in a 2011 interview. "\\
\bottomrule
\end{tabularx}
    \caption{Comparison of CS and FSD  on wikinews. We can see that Both FSD and CS can generate fluent sentences.  However, FSD is more sticky to the topic on these two prompts.}
    \label{tab:case_wikinews}
\end{table*}

\begin{table*}[ht]
\begin{tabularx}{\textwidth}{X}
\toprule
\small \textbf{Prompt-1:} \footnotesize <|endoftext|> With a satisfying protagonist completed, Sega turned to esteemed programmer Yuji Naka, who had impressed them through his work on Phantasy Star and the Genesis port \\
\midrule 
\small \textbf{FSD:} 
\footnotesize of Sonic the Hedgehog 2. Nana was a new face to the company, but he had a reputation for being a hard worker and a good communicator.  Nana's first task was to create a character that would be able to stand out from the crowd. He chose Sonic, a character who was already well known in Japan. The game's story was also designed to be unique, with the player controlling Sonic as he travels through a world that is not only different from any other Sonic game, but also from other games in the Sonic series.  The game was originally going to be called Sonic Adventure, but Sega decided to change the name to Sonic the Fighters.  Sega's decision to use a different name for the game was not a coincidence. In the early 1990s, the Japanese video game industry was in a state of flux. Sega was struggling to find its place in the market, and it was looking for a way to differentiate itself from Nintendo.  In the United States, Nintendo had been dominating the market for years, and Sega had struggled to gain a foothold. As a result, Sega began to look for ways to differentiate itself.  One of the ways Sega tried to do this was by creating a game that was\\
\midrule  
\small \textbf{CS:} \footnotesize of Final Fantasy. Naka was brought in to work on Sonic the Hedgehog, and the result was a game that had everything you could want in a Sonic game: speed, platforming, a sense of humor, and, of course, lots and lots of explosions.  Sonic 3 \& Knuckles is an incredibly well-made game, and the story is one of the best in the series to this day. It's not perfect, but it's a great example of what can be done with the Sonic formula.  In this article, we're going to take a look at some of the things that make Sonic 3 \& Knuckles such a great game, and how they were implemented in the game. If you're new to Sonic, this is a great place to start.  1. The Game Over screen is the best part of the game  The game over screen is where you get to see your score, how many lives you have left, and what level you're on. It's a great way to end a game, and it's one of the most memorable parts of the game.  2. There are a lot of secrets to find  The levels are packed with secrets, and they're hidden all over the\\
 
\midrule
\midrule 
\small \textbf{Prompt-2:} \footnotesize <|endoftext|> " Kiss You " was written by Kristoffer Fogelmark, Kristian Lundin, Albin Nedler, Savan Kotecha, Shellback \\
\midrule 
\small \textbf{FSD:} 
\footnotesize , and Jens-Christian Svenning. It was produced by the Swedish Film Institute and directed by Jonas Åkerlund.  The film was released in the United States on October 21, 2010.  Plot [ edit ]  A young man, who is a member of the "Kiss You" group, is on a date with a woman. He is very nervous and tries to hide his feelings. The woman, however, is not interested in him and leaves. The man is left alone and begins to cry.  Later, the man meets a man who has been in a coma for a year. The two men are friends and the man tells the young guy that he has a girlfriend. The young boy is shocked and asks the girl if she is his girlfriend. She says that she is not and that they are just friends.  The man then asks if he can kiss her. The girl says no and tells him that it is too soon. The man says that if it's not too late, he will kiss her.  The young couple then kiss and the woman says "I love you". The film ends with the two smiling.  Cast [ Edit ]  Production [ edit ];  Credits [ add\\
\midrule  
\small \textbf{CS:} 
\footnotesize .   Recorded and Produced at Soundworks in Gothenburg, Sweden. Mastered at Sterling Sound in Los Angeles, CA. Additional Recording at Studio G-A-N in Gothenburg.   Barcode: 7 27361 85928 2  Matrix / Runout (Variant 1): SON-A-1-SOUNDWAVES-BRAZIL-01-GOTHLAND-AUGUST-2014 (WEA) MADE IN JAPAN BY APPLE RECORDS A/S.  Matrix / Runout (Variant 2): SON-A-1-SOUNDWAVES-BRAZIL-01-GOTHLAND-AUGUST-2014 (WEA) MADE IN JAPAN BY APPLE RECORDS A/S.  Mastering SID Code (Variant 1): IFPI L574  Mastering SID Code (Variant 2): IFPI L574  Mould SID Code (Variant 1): IFPI 9085  Mould SID Code (Variant 2): IFPI 9085  Rights Society: GEMA\\

\bottomrule
\end{tabularx}
    \caption{Comparison of CS and FSD  on wikitext. CS obviously has a topic drift on Prompt-2.}
    \label{tab:case_wikitext}
\end{table*}

\begin{table*}[ht]
\begin{tabularx}{\textwidth}{X}
\toprule
\small \textbf{Prompt-1:} \footnotesize  The Sons vowed to combine their wisdoms and forsake their names and worship wholeheartedly the Unmarked Ones so that they one day might return. They would\\
\midrule 
\small \textbf{FSD:} \footnotesize be the first to bring the world back to the way it was before the fall.  The Firstborn  In the years that followed, the Sons of Svanir grew in power and influence. The First Born were the most powerful of their number, and they were known as the "Firstborn." They were also the most secretive, for they had no need of a public face. Their true name was kept secret, but they knew it well.  They were born of the same blood as their brothers, but their bodies were different. Instead of being born with a single eye, they had two. This was because the Firstborn were not born from the same mother as their brethren, but from a different mother.  This was not a problem for the FirstBorn, for their mother was a woman named Nithra. She was the daughter of an Unmarked One named Kord. Kords was an old man who had lived for many years. He had been a great warrior, and he had fought in many wars.  One day, Kord was attacked by a group of bandits. He was killed, and his body was taken to a nearby village. There, his wife, Nethra, was told of his death \\
\midrule  
\small \textbf{CS:} \footnotesize be a force of good in the world, and their god-kings would rule with an iron fist.  The Sons of Sanguinius had no need for this kind of power, but it was the only way they could keep from being wiped out by the Emperor's Children, who were the most powerful Space Marine Legion at the time. In the end, the Emperor's Children won the Great Crusade, and Sanguinius was forced to flee into the Eye of Terror, where he remained for a thousand standard years before being found by the Ultramarines.  Siege of Vraks  "You are a fool, Horus. I know what it is to be an outcast, and the shame is yours." — Fulgrim, to Horus during the Siege of Vraks (Unknown Date.M41)  In 998.M41, the Blood Angels were deployed to the world of Vraks, which had been the site of a bloody civil war between the Sons of Sanguinius and the Traitor Legion known as the World Eaters. The Sons of Sanguinius had taken the planet as a staging ground for their invasion of the Segmentum Obscurus, and were intent on using\\
 \midrule
 \midrule 

\small \textbf{Prompt-2:} \footnotesize But that's enough about aging, enough gloom and despondency. I shall tell you a story about when I myself was young and susceptible. It was in\\
\midrule 
\small \textbf{FSD:} \footnotesize the early days of the war, and I was a student at the University of Chicago. The war was raging, and we were all in a state of excitement. We were not yet aware of the fact that we had been born into a world of death.  I was studying in my room, with my books open on the table. Suddenly I heard a noise outside. I looked out the window, and saw a man running across the street. He was dressed in civilian clothes, but he was carrying a rifle.  "What are you doing?" I asked.  He stopped, looked at me, and then he ran back to his house.  The next day, I went to the university, where I found that the man had died.  It was a strange experience. For a moment I felt that I had lived in a dream. Then I remembered that it was only a dream.  But it is a fact, that in those days, when the world was full of fear, there were people who were prepared to die for their country.  And it is also a fact, that when we are young, we have a tendency to forget that there are people in the world who are prepared to sacrifice their lives for us\\
\midrule  
\small \textbf{CS:} 
\footnotesize the summer of 1884, and my family was on a visit to New York. We stayed at the Waldorf-Astoria, which was in the heart of Times Square. The hotel was a grand, opulent edifice, with marble floors, gold-trimmed walls, and a grand ballroom that could seat two thousand people at a time.  I was in the ballroom one day with my mother, father, and a few of my friends. It was a hot day, and the air was filled with steam from the boilers, which were in operation all over the hotel. We were sitting on the verandah, enjoying the cool breeze and the view of the city below. Suddenly there was a commotion in the ballroom, and a man in a black suit rushed in with a pistol in his hand. He pointed it at my mother, and demanded to know who she was. When she answered that she was my mother, the man drew his pistol and shot her point-blank in the chest. She fell to the floor, and I was left to look on in horror. My mother was dead, and there was nothing I could do to save her.  The next day, when I got home, I found a\\

\bottomrule
\end{tabularx}
    \caption{Comparison of CS and FSD  on book.}
    \label{tab:case_book}
\end{table*}

\begin{table*}[ht]
\begin{tabularx}{\textwidth}{X}
\toprule
\small \textbf{Prompt-1:} \footnotesize  A private security guard, defending the Panamanian based MV Almezaan, killed a Somali pirate who was attempting to hijack the ship on Tuesday.\\ 
\midrule 
\small \textbf{FSD:} \footnotesize The MV Maersk Alabama was sailing from the port of Bremerhaven in Germany to the Gulf of Aden when it was attacked by a pirate vessel.  The crew of the MV Ala'a were able to board the pirate ship and kill the Somali pirates.  "The security guards of the Almezan were on board the ship when they were attacked by pirates," said a statement from the German Maritime Ministry.  "The guards were armed with pistols and were able to shoot and injure the pirates."  The Almezan was able to return to Bremershaven and continue its journey to the gulf of Aden.  "We are grateful to the security forces of Germany and the United Arab Emirates for their quick response and for the safe return of the vessel," the statement said.  "This is a clear example of the cooperation between the maritime authorities of the two countries."  Piracy off the coast of Somalia has been on the rise in recent years.  In 2012, the UN Security Council passed a resolution to tackle the problem.  The resolution called for a "comprehensive and coordinated international response" to the problem.  It also called for the creation of a "global maritime force  \\
\midrule 
\small \textbf{FSD-vec:} \footnotesize  The incident occurred at about 10:30 a.m. local time (1:30 p.m. ET) when the crew of the MV Almezaan, a cargo vessel, spotted a small boat in the water off the coast of Somalia.  The security guard boarded the vessel and ordered the pirates to surrender.  "The pirate refused to comply with the order and fired a warning shot at the guard," said Captain John G. Smith, commander of the U.S. Navy's 7th Fleet.  "The armed guard then opened fire, killing the suspected pirate."  The Almezaan, which is owned by the Dutch company Maersk Line, was carrying about 1,000 tons of fuel oil, food and other supplies.  The Somali Pirates are known for their piracy and have been involved in several attacks on ships in the region.  In April, a Somali Pirate hijacked a tanker, the MSC Vos Hestia, off the Horn of Africa. The ship was eventually rescued by the Royal Navy.  Last year, a group of Somali pirates attacked a merchant shipping vessel, the SSV Punt, off the east coast of Africa.  The attack resulted in the\\
\midrule
\midrule 
\small \textbf{Prompt-2:} \footnotesize The Space Shuttle Discovery, flying the STS-133 mission, has successfully rendezvoused and docked with the International Space Station (ISS) today at \\
\midrule 
\small \textbf{FSD:} \footnotesize 7:30 a.m. EDT.  The docking was completed at 8:15 a.m.  "The crew of STS 133 is safely back on Earth," said NASA Administrator Charles Bolden. "This is a great day for NASA and the nation. We are grateful to the crew and their families for their dedication and professionalism."  "This is a very exciting day for the entire NASA family," added NASA Chief Technologist David Weaver. "The shuttle program has been a tremendous success, and we are proud of the work of the men and women of the shuttle fleet. This is a day of celebration for all of us at NASA."  The STS flight was the first of two planned shuttle missions to the ISS. The second mission, STS-134, is scheduled for launch in late November.  "Today's successful docking marks the end of a long journey for the space shuttle," noted NASA Administrator Boldin. "It is a testament to the hard work and dedication of the astronauts and engineers who have flown on these missions. It is also a tribute to the strength of the American spirit and the ingenuity of our nation's scientists and engineers."  "We are very proud of the team that has worked so hard \\
\midrule 
\small \textbf{FSD-vec:} \footnotesize 7:30 a.m. EDT.  The docking was completed at 8:15 a.m. EDT, NASA officials said.  "This is a great day for all of us," said Rick Mastracchio, commander of the space shuttle program. "We are very proud of the crew and the team that worked so hard to get this done."  The shuttle's arrival at the ISS marks the first time a U.S. spacecraft has docked with the orbiting laboratory since the end of the Apollo era in 1972.  "Today is a historic moment for the United States and the world, and we are grateful to the astronauts and their families for their dedication and professionalism throughout this process," President Obama said in a statement.  "I want to thank the entire NASA team for their efforts to bring the Space Shuttle back to Earth safely, and I congratulate the STS-133 crew on their successful return to flight. This is a very special achievement for our nation, and it is a testament to the ingenuity and determination of the American people. We are all looking forward to seeing what they do next!"  The mission, which was originally scheduled to launch on May 20, will be the last of the current generation of shutt\\
\bottomrule
\end{tabularx}
    \caption{Comparison of FSD and FSD-vec on wikinews.}
    \label{tab:case_wikinews_ours}
\end{table*}

\begin{table*}[ht]
\begin{tabularx}{\textwidth}{X}
\toprule
\small \textbf{Prompt-1:} \footnotesize <|endoftext|> A tropical depression developed near Yap on November 11. The system intensified gradually as it tracked quickly westward toward the Philippines. An anticyclone aloft \\ 
\midrule 
\small \textbf{FSD:} \footnotesize over the central Pacific was responsible for the development of the system.  The system was located about 1,000 miles west of Yape, Papua New Guinea, and about 2,500 miles east of Guam. It was moving west at about 15 mph.  A tropical storm warning was issued for Yap and the surrounding area.  On November 12, the system strengthened to a tropical cyclone.  It was named Yolanda after the city in the Philippines where it formed.  Yoland, a small town in Yapto, is located on the northern coast of the island of Yap.  The National Hurricane Center said the storm was expected to strengthen further and become a hurricane by the time it reaches the western Pacific.  Forecasters said Yolan will likely become a Category 1 hurricane on Friday.  "The center of this system is expected to move over land near the southern Philippines on Saturday and Sunday," the center said. "The maximum sustained winds are near 45 mph, with higher gusts."  Yolanda is about 3,200 miles south of Manila.  "Yolan is forecast to become a major hurricane later this week," the National Weather Service said. "It is likely \\
\midrule 
\small \textbf{FSD-vec:} \footnotesize over the central Pacific was responsible for the development of the storm.  The typhoon, which is named after the island of Yape, formed in the western part of the Central Philippine Sea. It moved northwest at a speed of about 10 mph (16 km/h) and then turned east-northeast.  It weakened slightly during the night, but remained a Category 1 hurricane with maximum sustained winds of 115 miles per hour.  Yap is located about 2,000 miles southwest of Manila.  A Tropical Storm Warning was issued for the northernmost parts of the country, including the provinces of Cebu, Iloilo, and Bohol.  Forecasters said the cyclone could bring heavy rain and strong winds to the southern and eastern portions of the archipelago.  "We are expecting some very intense rainfall, especially in the south and southeast," National Disaster Risk Reduction and Management Council (NDRRMC) Director General Joel Pagdilao told ABS-CBN News.  "There will be gusty wind, so we need to be prepared for that."  Pagdilao added that the NDRRMC is monitoring the situation closely.  "Our teams are on \\
\midrule
\midrule 
\small \textbf{Prompt-2:} \footnotesize <|endoftext|> Malaysia LNG is a Liquefied Natural Gas ( LNG ) manufacturing complex located in Bintulu that currently contains eight LNG trains with a ninth \\
\midrule 
\small \textbf{FSD:} \footnotesize  train under construction. The complex is owned by the Malaysian government and is operated by Malaysia's state-owned Petronas.  The complex was built in the early 1990s to supply LPG to the domestic market. In 2004, Petrolio, a subsidiary of Petronas, acquired the LNGL project and began to produce LNB.  In 2009, the government of Malaysia announced that it would be building a second LGNL complex in Kuala Lumpur. This second complex, which is currently under development, will be capable of producing up to 1.5 million tonnes of LBN per year.  Petronas has stated that the second facility will have a capacity of up to 2.2 million tons per annum.  LNG train  The first LNG train was delivered to BINTUL in April 2010. The train is capable of transporting up to 4.3 million litres of LNB per day.  A second train, which was originally scheduled to be delivered in 2012, was delayed due to a number of issues.  In September 2012, Petromonas announced that the first train would be completed by December 2012. However, in January 2013, the company announced that the train would not be ready\\
\midrule 
\small \textbf{FSD-vec:} \footnotesize train under construction. The facility is owned by Malaysia's state-owned Petronas.  The plant was built in the early 1990s and has been used to produce liquified natural gas for export. It is one of the largest industrial facilities in Southeast Asia.  In addition to the four existing rail lines, there are plans to build a fifth line.  Liquifed-natural gasses are produced from coal, oil, or biomass. They are then transported by pipeline to the terminal, where they are loaded onto tankers and shipped to other countries.  Malaysia produces about 1.5 million tonnes of liquid fuel per year.  A new railway will be constructed to transport the gas to the port.  Construction of the new train began in 2009.  It is expected to be completed by 2015.  The first two tracks were laid in 2010.  The third track was installed in 2011.  The fourth and final section of the railway was opened in 2012.  The LNGL project is part of the Malaysian government's plan to increase its energy security.  Petrol prices have been rising steadily since the beginning of the year.  According to the Petroleum Ministry, petrol costs\\
\bottomrule
\end{tabularx}
    \caption{Comparison of FSD and FSD-vec on wikitext.}
    \label{tab:case_wikitext_ours}
\end{table*}

\begin{table*}[ht]
\begin{tabularx}{\textwidth}{X}
\toprule
\small \textbf{Prompt-1:} \footnotesize The person slowly backed out of the snag. Olivia was expecting either a kid or a deranged hermit, so she was startled to see a young woman, \\ 
\midrule 
\small \textbf{FSD:} \footnotesize  maybe in her early twenties, standing there. She was dressed in a black dress, with a white blouse and black pants. Her hair was tied back in a ponytail, and she had a small, round face.  "Hello, Olivia," the woman said. "I'm Sarah. I'm here to help you."  Olivia's eyes widened. "You're Sarah?"  Sarah nodded. "Yes. I've been looking for you for a while. You're the one who saved my life."  She smiled. "Thank you. It was my pleasure."  Oblivious to the fact that she'd just met a stranger, Olivia walked over to Sarah and hugged her. Sarah returned the hug, and Olivia felt her heart swell with gratitude.  "I've got to go," Sarah said.  As Olivia stood up, she noticed that Sarah had left her purse behind.  "Where's your purse?" Olivia asked.  Sarah shrugged. "It's in my room. I'll be back soon."  Olivetti  The next day, Sarah came back to Olivia's room.  "You look tired," she said.  I know, I\\
\midrule 
\small \textbf{FSD-vec:} \footnotesize   maybe in her early twenties, standing there. She had long, dark hair, and wore a black dress with a white collar. Her eyes were closed, but they looked like those of a child.  "Hello," said the girl. "I'm Emily. I've been waiting for you."  Olivia's heart sank.  Emily smiled.  She took off her glasses and wiped her eyes.  The two of them walked into the house.  They sat down on the couch.  It was quiet.  Olivia felt as if she'd been transported back to that night.  But this time, it wasn't just because of the silence.  There was something else.  Something different.  The air was warm.  And, somehow, there was a sense of peace.  Olivia couldn't quite put her finger on it.  Maybe it was the fact that the room was empty.  Or maybe, just maybe, it was because there was someone sitting across from her.  A tall, thin man.  He didn't look at her.  Instead, he stared at the floor.  His face was expression\\
\midrule
\midrule 
\small \textbf{Prompt-2:} \footnotesize The trip ended far too soon; I was a bit confused as we pulled up to a small beach cottage on the far side of the inlet. The whit \\
\midrule 
\small \textbf{FSD:} \footnotesize ewashed walls were covered in graffiti, and the windows were boarded up. I asked the owner if he knew where we could find a place to stay, but he didn't seem to know either. We decided to walk back to the car and head back out to the inlets.  We walked along the beach for a while, until we came across a large group of people sitting on a bench. They were all dressed in white, and they were talking about something. It was hard to make out what they were saying, but I could tell they had a lot of energy.  I asked one of the guys if I could join them, and he said sure. He led me to a spot on the beach where he had set up a tent. As we sat down, I noticed that the guy was wearing a white shirt with a black tie.  "What's your name?" I asked.  "I'm John," he replied.  He was tall, with dark hair and a beard. He was also wearing jeans and a t-shirt.  John was an older man, probably in his late 50s or early 60s. He had dark skin, a thin mustache, and was balding.  As we \\
\midrule 
\small \textbf{FSD-vec:} \footnotesize ewashed walls were covered in graffiti, and the windows were boarded up. We walked inside, and found a large, empty room.  "I'm sorry, but there's nothing here," said my guide.  We sat down at the table, and he told us about his life. He had been born in Mexico, but moved to the United States when he was five. His parents divorced, and he lived with his mother.  He worked as a cook, and then became a security guard. When that job dried up, he started working for a construction company.  When the recession hit, things got worse.  The economy was in shambles, and many people lost their jobs.  His boss fired him, and sent him to live with his grandmother.  That's when everything changed.  A few months later, his father died.  It was hard for him to cope.  So, after living with his grandparents for a while, he decided to move back to Mexico.  But, before he left, he took some photos.  One day, while walking home, he saw a man who looked like him.  They talked, and eventually, they agreed to meet\\
\bottomrule
\end{tabularx}
    \caption{Comparison of FSD and FSD-vec on book.}
    \label{tab:case_book_ours}
\end{table*}


\end{document}